\documentclass[9pt,a4paper,reqno]{article}

\usepackage{times}
\usepackage[left=0.5in,right=0.5in,top=0.6in,bottom=0.8in]{geometry} 
\usepackage{multicol}
\usepackage{epstopdf}
\usepackage{multicol}
\usepackage{authblk}
\usepackage{array}
\usepackage{graphicx}
\usepackage{caption}
\usepackage{listings}
\usepackage{tikz}
\usepackage{subcaption}
\usepackage[T1]{fontenc}
\usepackage{float}
\usepackage{amssymb,amsfonts,amssymb,amsmath}
\usepackage{xcolor}

\usepackage{algorithm}
\usepackage[noend]{algpseudocode}
\usepackage{titling}

\usepackage{etoolbox} %

\usepackage[backend=bibtex, style=ieee, sorting=none]{biblatex} %
\usepackage[british]{babel}
\usepackage{csquotes}
\usetikzlibrary{patterns}
\usetikzlibrary{arrows}
\usetikzlibrary{shapes}
\newenvironment{natabstract}{%
\begin{quote} \bf}
{\end{quote}}

\newcounter{lastnote}

\def\d{\text{d}\,}
\def\R{\mathbb{R}}
\def\N{\mathbb{N}}
\def\C{\mathbb{C}}
\def\Om{\Omega}
\def\la{\lambda}

\DeclareMathOperator*{\minimize}{minimize}

\DeclareMathOperator{\argmax}{arg\!\max}

\makeatletter
\newenvironment{mytitlepage}%
  {\def\@thanks{}}%
  {}
\makeatother

\title{On instabilities of deep learning in image reconstruction - Does \\AI come at a cost?}  
\vspace{-5mm}

\newcommand*\samethanks[1][\value{footnote}]{\footnotemark[#1]}
\author{Vegard Antun\thanks{Department of Mathematics, University of Oslo, Norway}, 
        \and Francesco Renna\thanks{Instituto de Telecomunica\c{c}\~{o}es, 
                                    Faculdade de Ci\^{e}ncias da Universidade do Porto, Portugal}, 
        \and Clarice Poon\thanks{Department of Mathematical Sciences, 
                                 University of Bath, UK}, 
        \and Ben Adcock\thanks{Department of Mathematics, Simon Fraser University, Canda}, 
        \and Anders C. Hansen*\thanks{Department of Applied Mathematics and 
                                     Theoretical Physics (DAMPT), 
                                     University of Cambridge, UK (* corresponding author)}
                                     $^{,}$\samethanks[1]}

\date{}

\thanksmarkseries{arabic}

\begin{document}

\begin{mytitlepage}
\maketitle 

\end{mytitlepage}

\vspace{-14mm}
\begin{natabstract}
\linespread{0}
Deep learning, due to its unprecedented success in tasks such as image classification, has emerged as a new tool in image reconstruction with potential to change the field. In this paper we demonstrate a crucial phenomenon: deep learning typically yields unstable
methods for image reconstruction. The instabilities usually occur in several forms: (1) tiny, almost undetectable perturbations, both in the image and sampling domain, may result in severe artefacts in the reconstruction, (2) a small structural change, for example a tumour, may not be captured in the reconstructed image and (3) (a counterintuitive type of instability) more samples may yield poorer performance.  Our new stability test with algorithms and easy to use software detects the instability phenomena. The test is aimed at researchers to test their networks for instabilities and for government agencies, such as the Food and Drug Administration (FDA), to secure safe use of deep learning methods. 
\end{natabstract}

\begin{multicols}{2}
\setcounter{footnote}{0}
\linespread{0.9}
\renewcommand{\thefootnote}{\fnsymbol{footnote}}
\fontsize{10}{12}
\selectfont
\newrefsegment

The importance of stable and accurate methods for image reconstruction for inverse problems is hard to over estimate. These techniques form the foundation for essential tools across the physical and life sciences such as Magnetic Resonance Imaging (MRI), Computerised Tomography (CT), fluorescence microscopy, electron tomography, Nuclear Magnetic Resonance (NMR), radio interferometry, lensless cameras etc. Moreover, stability is traditionally considered a necessity in order to secure reliable and trustworthy methods used in, for example, cancer diagnosis. Hence, there is an extensive literature on designing stable methods for image reconstruction in inverse problems \cite{Nature_inverse, Candes_PNAS, stability_book, PCHansen}.

Artificial intelligence (AI) techniques such as deep learning and neural networks \cite{LeCun_Nature} have provided a new paradigm with new techniques in inverse problems \cite{Rosen_Nature, Nature_highlights,  McCann17, jin17, DeepMRI2d, DAGAN, Hammernik18, Lucas18, Mardani18} that may change the field. In particular, the reconstruction algorithms learn how to best do the reconstruction based on training from previous data, and through this training procedure aim to optimise the quality of the reconstruction. This is a radical change from the current state of the art both from an engineering, physical and mathematical point of view. 

AI and deep learning has already changed the field of computer vision and image classification \cite{Elad, Girshick14, Alexnet12,Zhou14}, where the performance is now referred to as super human \cite{he2015delving}. However, the success comes with a price. Indeed, the methods are highly unstable. It is now well established \cite{kanbak17, moosavi16,
moosavi17, szegedy13,Fawzi17} that high performance deep learning methods for image classification are subject to failure given tiny, almost invisible perturbation of the image. An image of a cat may be classified correctly, however, a tiny change, invisible to the human eye, may cause the algorithm to change its classification label from cat to a fire truck, or another label far from the original.   

In this paper we establish the instability phenomenon of deep learning in image reconstruction for inverse problems. A potential surprising conclusion is that the phenomenon may be independent of the underlying mathematical model. For example, MRI is based on sampling the Fourier transform whereas CT is based on sampling the Radon transform. These are rather different models, yet the instability phenomena happen for both sampling modalities when using deep learning.

\begin{figure*}[t]
    \centering
    \vspace{-1cm}
    \begin{tabular}{m{0.175\textwidth}m{0.175\textwidth}m{0.175\textwidth}m{0.175\textwidth}m{0.175\textwidth}}
    $\qquad \quad \text{Original }  |x|$ & $\quad\quad\quad\quad |x + r_1|$ & $\quad\quad\quad\quad |x + r_2|$& $\quad\quad\quad\quad  |x + r_3|$& $\quad\quad$   SoA from $Ax$\\

     \includegraphics[width=0.19\textwidth]{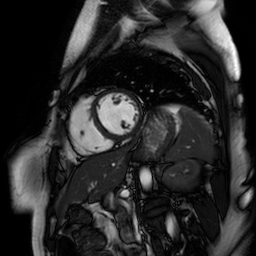}
    &\includegraphics[width=0.19\textwidth]{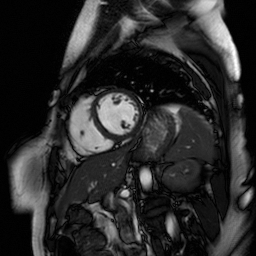}
    &\includegraphics[width=0.19\textwidth]{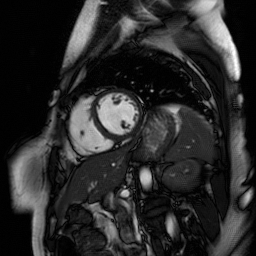}
    &\includegraphics[width=0.19\textwidth]{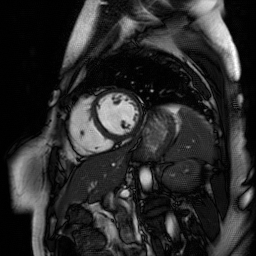}
    &\includegraphics[width=0.19\textwidth]{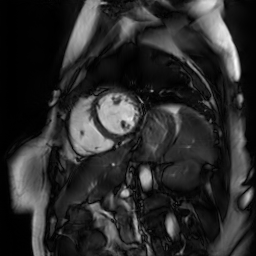}\\
 
  $\text{Deep MRI (DM)} \, f(Ax)$ & $\quad \text{DM } \, f(A(x + r_1))$ & $\quad \text{DM } \, f(A(x + r_2))$ & $\quad \text{DM } \, f(A(x + r_3))$  & $\quad$SoA from $A(x+r_3)$\\

     \includegraphics[width=0.19\textwidth]{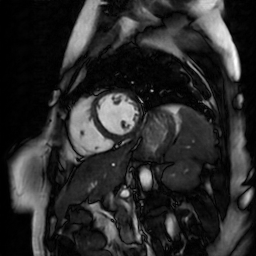}
    &\includegraphics[width=0.19\textwidth]{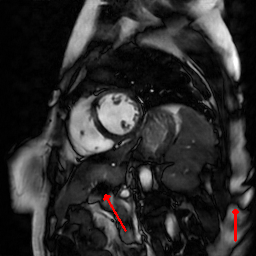}
    &\includegraphics[width=0.19\textwidth]{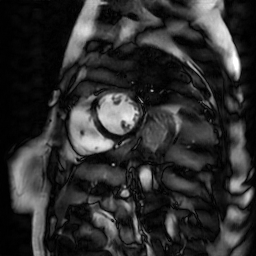}
    &\includegraphics[width=0.19\textwidth]{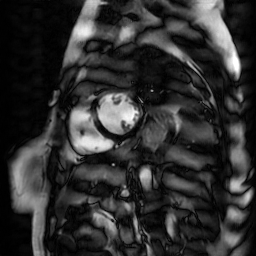}
    &\includegraphics[width=0.19\textwidth]{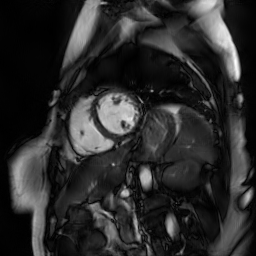}\\

    \end{tabular}
    \caption{\label{fig:deep_mri_pert11} Perturbations $r_j$ with $|r_1| < |r_2| < |r_3|$ are added to the image $x$. Upper row images (1)-(4): original image $x$ and perturbations $x+r_j$. Lower row images (1)-(4) reconstruction from $A(x+r_j)$ using the Deep MRI network $f$, where $A$ is a subsampled Fourier transform (33\% subsampling), see Methods and SI for details. Rightmost column: reconstruction from $Ax$ and $A(x+r_3)$ using a state-of-the-art (SoA) method, see Methods for details. Note how the artefacts (red arrows) are hard to dismiss as non-physical. 
}
\vspace{-4mm}
\end{figure*}

There is, however, a big difference between the instabilities of deep learning for image classification and our results on instabilities of deep learning for image reconstruction. Firstly, in the former case there is only one thing that could go wrong: a small perturbation results in a wrong classification. In image reconstruction there are several potential forms of instabilities. In particular, we consider three crucial issues: (1) {\it instabilities with respect to tiny perturbations}, (2) {\it instabilities with respect to small structural changes} (for example a brain image with or without a small tumour), (3) {\it instabilities with respect to changes in the number of samples}. Secondly, the two problems are totally unrelated. Indeed, the former problem is, in its simplest form, a decision problem, and hence the decision function ("is there a cat in the image?") to be approximated is necessarily discontinuous.  However, the problem of reconstructing an image from Fourier coefficients, as is the problem in MRI, is completely different. In this case there exist stable and accurate methods that depend continuously on the input. It is therefore paradoxical that deep learning leads to unstable methods for problems that can be solved accurately in a stable way.  

The networks we have tested are unstable either in the form of (1) or (2) or both. Moreover, networks that are highly stable in one of the categories tend to be highly unstable in the other. The instability in form of (3), however, occur for some networks but not all.     
The new findings raise two fundamental questions:

\vspace{1mm}
(i) Does AI, as we know it, come at a cost? Is instability a necessary by-product of our current AI techniques?

\vspace{0.5mm}
(ii) Can reconstruction methods based on deep learning always be safely used in the physical and life sciences?  Or, are there cases for which instabilities may lead to, for example, incorrect medical diagnosis if applied in medical imaging? 
\vspace{1mm}

The scope of this paper is on the second question, as the first question is on foundations, and our stability test provides the starting point for answering question (ii). However, even if instabilities occur, this should not rule out the use of deep learning methods in inverse problems. In fact, one may be able to show, with large empirical statistical tests, that the artefacts caused by instabilities occur infrequently. As our test reveals, there is a myriad of different artefacts that may occur, as a result of the instabilities, suggesting vast efforts needed to answer (ii). 
A detailed account is in the conclusion. 

\begin{figure*}[htbp]
    \centering
    \vspace{-1cm}
    \begin{tabular}{m{0.175\textwidth}m{0.175\textwidth}m{0.175\textwidth}m{0.175\textwidth}m{0.175\textwidth}}   
    $\qquad \quad \text{Original }  x$ & $\quad\quad\quad\quad |x + r_1|$ & $\quad\quad\quad\quad |x + r_2|$& $\quad\quad\quad\quad  |x + r_3|$& $\quad\quad\quad\quad   |x + r_4|$\\

     \includegraphics[width=0.19\textwidth]{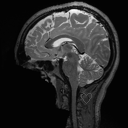}
    &\includegraphics[width=0.19\textwidth]{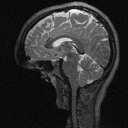}
    &\includegraphics[width=0.19\textwidth]{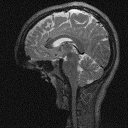}
    &\includegraphics[width=0.19\textwidth]{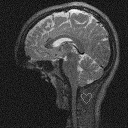}
    &\includegraphics[width=0.19\textwidth]{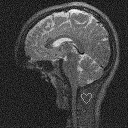}\\
 
  $\text{AUTOMAP (AM)} \, f(Ax)$ & $\quad \text{AM} \, f(A(x + r_1))$ & $\quad \text{AM}\, f(A(x + r_2))$ & $\quad \text{AM} \, f(A(x + r_3))$ & $\quad \text{AM} \, f(A(x + r_4))$\\

     \includegraphics[width=0.19\textwidth]{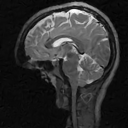}
    &\includegraphics[width=0.19\textwidth]{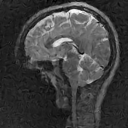}
    &\includegraphics[width=0.19\textwidth]{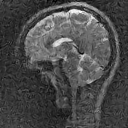}
    &\includegraphics[width=0.19\textwidth]{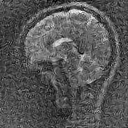}
    &\includegraphics[width=0.19\textwidth]{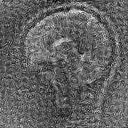}\\
    
      SoA from $A(x)$ &  SoA from $A(x+r_1)$ &  SoA from $A(x+r_2)$ &  SoA from $A(x+r_3)$ &  SoA from $A(x+r_4)$\\

     \includegraphics[width=0.19\textwidth]{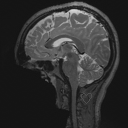}
    &\includegraphics[width=0.19\textwidth]{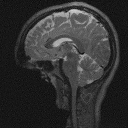}
    &\includegraphics[width=0.19\textwidth]{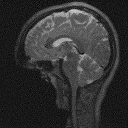}
    &\includegraphics[width=0.19\textwidth]{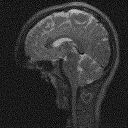}
    &\includegraphics[width=0.19\textwidth]{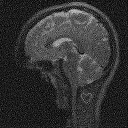}\\

    \end{tabular}
    \caption{\label{fig:automap_pert} Perturbations $\tilde r_j$ are added to the measurements $y = Ax$, where $|\tilde r_1| < |\tilde r_2| < |\tilde r_3| < |\tilde r_4|$ and $A$ is a subsampled Fourier transform (60\% subsampling). To visualise we show $|x+r_j|$ where $y+\tilde r_j = A(x+r_j)$. Upper row: original image $x$ with perturbations $r_j$. Middle row: reconstructions from $A(x + r_j)$ by the AUTOMAP network $f$. Lower row: reconstructions from $A(x + r_j)$ by a state of the art method (see Methods for details). A detail in form of a heart, with varying intensity,  is added to visualise the loss in quality. 
}
\vspace{-4mm}
\end{figure*}

\vspace{-2mm}

\subsubsection*{The instability test}

\vspace{-2mm}
The instability test is based on the three instability issues mentioned above. We consider instabilities with respect to the following:

\textbf{Tiny perturbations.} The tiny perturbation could be in the image domain or in the sampling domain. When considering medical imaging, a perturbation in the image domain could come from a slight movement of the patient, small anatomic differences between people etc. The perturbation in the sampling domain may be caused by malfunctioning of the equipment or the inevitable noise dictated by the physical model of the scanning machine. However, a perturbation in the image domain may imply a perturbation in the sampling domain. Also, in many cases, the mathematical model of the sampling reveals that such a sampling process implies an operator that is surjective onto its range, and hence there exists a perturbation in the image domain corresponding to the perturbation in the sampling domain. Thus, a combination of all these factors may yield perturbations that in a worst case scenario may be quite specific, hard to model and hard to protect against unless one has a completely stable neural network.

The instability test includes algorithms that do the following. Given an image and a neural network, designed for image reconstruction from samples provided by a specific sampling modality, the algorithm searches for a perturbation of the image that makes the most severe change in the output of the network while still keeping the perturbation small. In a simple mathematical form this can be described as follows. Given an image $x \in \mathbb{R}^N$ (we interpret an image as a vector for simplicity), a matrix $A \in \mathbb{C}^{m \times N}$ representing the sampling modality (for example a discrete Fourier transform modelling MRI) and a neural network $f: \mathbb{C}^m \rightarrow \mathbb{C}^N$, the neural network reconstructs an approximation $\tilde x$ to $x$ defined by
$
y = Ax, 
$
where 
$
\tilde x = f(y).
$
The algorithm seeks an $r \in \mathbb{R}^N$ such that 
\[
\|f(y + Ar) - f(y)\| \text{ is large, while } \|r\| \text{ is small,} 
\]   
see the methods section for details. 
However, the perturbation could, of course, be put on the measurement vector $y$ instead.
    
\textbf{Small structural changes in the image.} By structural change we mean a change in the image domain that may not be tiny, and typically significant and clearly visible, however still small (for example a small tumour). The purpose is to check if the network can recover important details that are crucial in, for example, medical assessments. In particular, given the image $x \in \mathbb{R}^N$ we add a perturbation $r \in \mathbb{R}^N$, where $r$ is a detail that is clearly visible in the perturbed image $x+r$, and check if $r$ is still clearly visible in the reconstructed image $\hat x = f(A(x+r))$.  
In this paper we consider the symbols from cards as well as letters. In particular, we add the symbols
$
\spadesuit, \heartsuit, \diamondsuit, \clubsuit
$ and the letters 
$\text{ CAN U SEE IT}$
to the image. The reason for this is that card symbols as well as letters are fine details that are hard to detect, and thus represent a reasonable challenge for the network. If the network is able to recover these small structural changes it is likely to recover other details of the same size. On the other hand, if the network fails on these basic changes, it is likely to fail on other details as well. The symbols can, of course, be specified depending on the specific application.  Our choice is merely for illustration. 

{\it Important note:} When testing stability, both with respect to tiny perturbations and with respect to small structural changes, the test is always done in comparison with a state-of-the-art (SoA abbreviated) stable method in order to check that any instabilities produced by the neural network is due to the network itself and not because of ill-conditioning of the inverse problem. The state-of-the-art methods used are based on compressed sensing and sparse regularisation \cite{CandesRombergTao, donohoCS, Rudin}. These methods often come with mathematical stability guaranties \cite{SMSparsity}, and are hence suitable as benchmarks (see the Methods for details).

\textbf{Changing the number of samples in the sampling device (such as the MRI or CT scanner)}. Typical state-of-the-art methods share a common quality; more samples imply better quality of the reconstruction. Given that deep learning neural networks in inverse problems are trained given a specific sampling pattern, the question is how robust is the trained network with respect to changes in the sampling. The test checks whether the quality of the reconstruction deteriorates with more samples. This is a crucial question in applications. For example the recent implementation of compressed sensing on Philips MRI machines allows the user to change the under sampling ration for every scan. This means that if a network is trained on $25\%$ subsampling, say, and suddenly the user changed the subsampling ratio to $35\%$ one would want an improved recovery. 
If the quality deteriorates or stagnates with more samples, this means that one will have to produce networks trained for each and every combination of subsampling that the machine allows for. Finally, due to the other instability issues, every such network must individually be empirically statistically tested to detect whether the occurrence of instabilities is rare or not. It is not enough to test on only one neural network, as their instabilities may be completely different.

\vspace{-2mm}

 \subsubsection*{Testing the test}
 
 \vspace{-2mm}
 
 We test six deep learning neural networks selected based on their strong performance, wide range in  architectures, difference in sampling patterns and subsampling ratios, as well as their difference in training data. The specific details about the architecture and the training data of the tested networks can be found in the supplementary information (SI).

\begin{figure*}[t]
    \centering
    \vspace{-1cm}
    \begin{tabular}{m{0.22\textwidth}m{0.22\textwidth}m{0.22\textwidth}m{0.22\textwidth}}
    $\qquad \qquad \text{Original }  x$ & $\quad\quad\quad\quad\quad   x + r_1$ & $\qquad \qquad \text{Original }  \tilde x$& 
    $\quad\quad\quad\quad\quad  \tilde x + r_2$\\
    \includegraphics[width=0.227\textwidth]{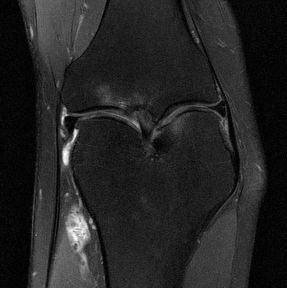} 
    &\includegraphics[width=0.227\textwidth]{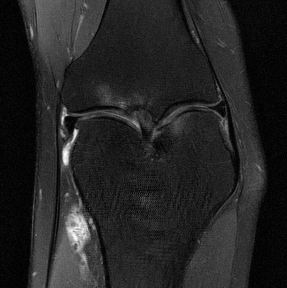} 
    &\includegraphics[width=0.23\textwidth]{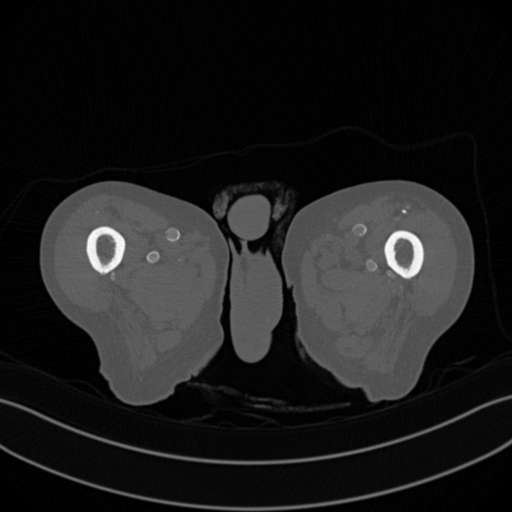} 
    &\includegraphics[width=0.23\textwidth]{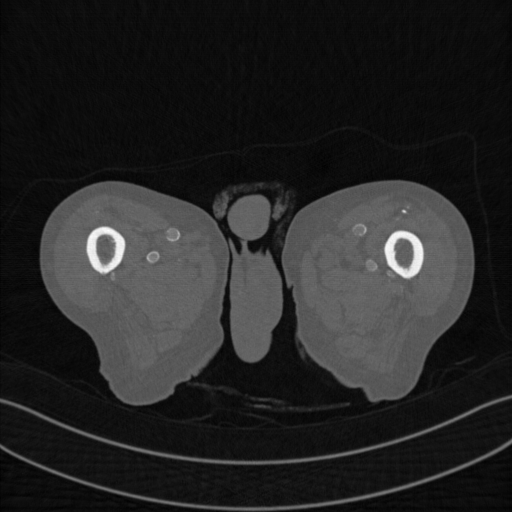} \\
    
     $\qquad \quad \text{MRI-VN} \, f(Ax)$ & $\quad \text{MRI-VN} \, f(A(x + r_1))$ & $\qquad \quad \text{MED 50} \, \tilde f(\tilde A\tilde x)$& 
    $\quad \text{MED 50} \, \tilde f(\tilde A(\tilde x+r_2))$\\
    \includegraphics[width=0.227\textwidth]{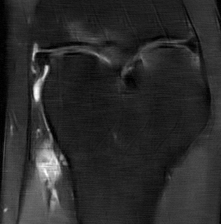} 
    &\includegraphics[width=0.227\textwidth]{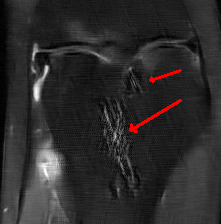} 
    &\includegraphics[width=0.23\textwidth]{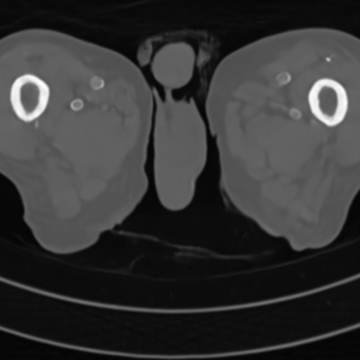} 
    &\includegraphics[width=0.23\textwidth]{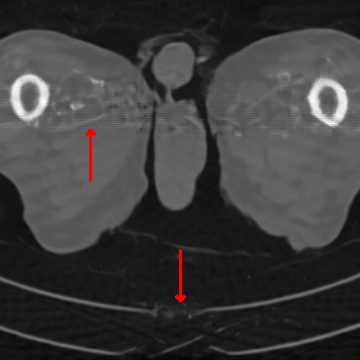} \\
    
    $\qquad \quad \text{SoA from } Ax$ & $\qquad \text{SoA from } A(x+r_1)$ & $\qquad \quad \text{SoA from } \tilde A\tilde x$& 
    $\qquad \text{SoA from }\tilde A(\tilde x+r_2)$\\
    \includegraphics[width=0.227\textwidth]{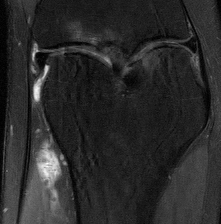} 
    &\includegraphics[width=0.227\textwidth]{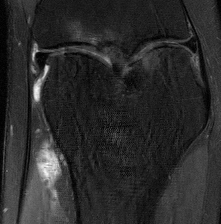} 
    &\includegraphics[width=0.23\textwidth]{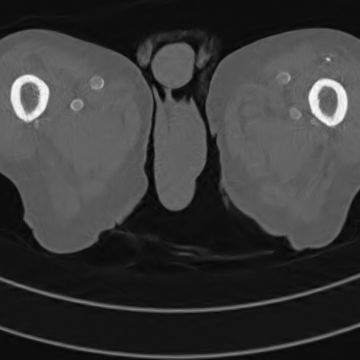} 
    &\includegraphics[width=0.23\textwidth]{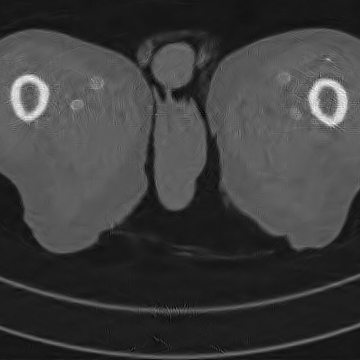} \\

  \end{tabular}
    \caption{\label{fig:deep_mri_pert2} Perturbations $r_1,r_2$ are added to the images $x$ and $\tilde x$ in the first row. The reconstructions by the network $f$ (MRI-VN), from $Ax$ and $A(x+r_1)$, and the network $\tilde f$ (MED 50), from $\tilde A\tilde x$ and $\tilde A(\tilde x +r_2)$ are shown in the second row. $A$ is a subsampled discrete Fourier transform and $\tilde A$ is a subsampled Radon transform. State-of-the-art (SoA) comparisons are shown in the last row. 
 }
 \vspace{-4mm}
\end{figure*}

  {\it Important note:} The tests performed are not designed to test deep learning against state-of-the-art in terms of performance on specific images. The test is designed to detect the instability phenomenon. Hence, the comparison with state-of-the-art is only to verify that the instabilities are exclusive only to neural networks based on deep learning, and not due to an ill-conditioning of the problem itself. Moreover, as is clear from the images, in the unperturbed cases, the best performance varies between neural networks and state-of-the-art. The list of networks is as follows:

 {\it AUTOMAP} \cite{Rosen_Nature}:  This is a neural network for low resolution single coil MRI with 60\% subsampling.  The training set consists of brain images with added white noise to the Fourier samples.  
 
{\it DAGAN} \cite{DAGAN}: This network is for medium resolution single coil MRI with 20\% subsampling, and is trained with a variety of brain images.  

{\it Deep MRI} \cite{DeepMRI2d}: This neural network is for medium resolution single coil MRI with 33\% subsampling. It is trained with detailed cardiac MR images.

 {\it Ell 50} \cite{jin17}: Ell 50 is a network for CT or any Radon transform based inverse problem. It is trained on images containing solely ellipses (hence the name Ell 50). The number 50 refers to the number of lines used in the sampling in the sinogram. 
 
 {\it Med 50} \cite{jin17}: Med 50 has exactly the same architecture as Ell 50 and is used for CT, however, it is trained with medical images (hence the name Med 50) from the Mayo Clinic database. The number of lines used in the sampling from the sinogram is 50.
 
{\it MRI-VN} \cite{Hammernik18}: This network is for medium to high resolution parallel MRI with 15 coil elements and 15\% subsampling. The training is done with a variety of knee images.

\vspace{-2mm}

\subsubsection*{Stability with respect to tiny perturbations}

\vspace{-2mm}

Below follows the description of the test applied to some of the networks where we detect instabilities with respect to tiny perturbations.

\begin{figure*}[htbp]
    \centering
    \vspace{-1cm}
    \begin{tabular}{m{0.22\textwidth}m{0.22\textwidth}m{0.22\textwidth}m{0.22\textwidth}}
      $\qquad \quad \text{Original } x_1 + r_1$ & $\, \, \, \text{Original } x_1 + r_1 \, \text{(zoomed)}$ & $\quad\quad \text{Ell 50} \, f_1(A_1x_1 + r_1)$& 
    $\quad \,\, \text{SoA from } A_1(x_1+r_1)$\\
    \includegraphics[width=0.227\textwidth]{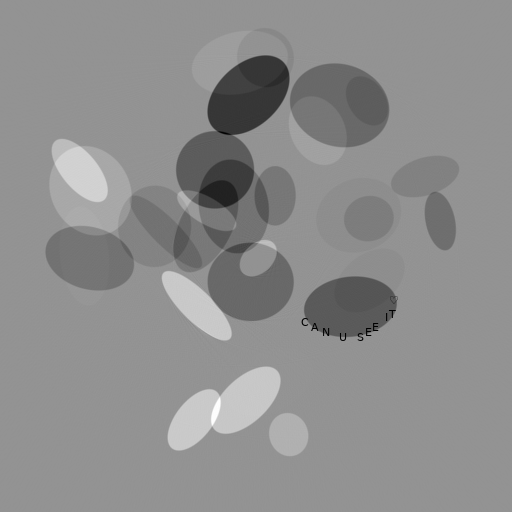} 
    &\includegraphics[width=0.227\textwidth]{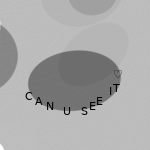} 
    &\includegraphics[width=0.23\textwidth]{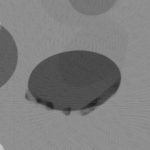} 
    &\includegraphics[width=0.23\textwidth]{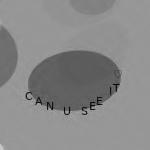} \\
    
     $\qquad \quad \text{Original } x_2 + r_2$ & $\, \, \, \text{Original } x_2 + r_2 \, \text{(zoomed)}$ & $\quad\quad \text{DAGAN} \, f_2(A_2x_2 + r_2)$& 
    $\quad \,\, \text{SoA from } A_2(x_2+r_2)$\\
    \includegraphics[width=0.227\textwidth]{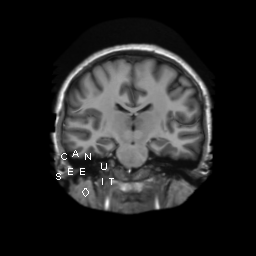} 
    &\includegraphics[width=0.227\textwidth]{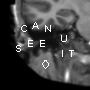} 
    &\includegraphics[width=0.23\textwidth]{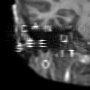} 
    &\includegraphics[width=0.23\textwidth]{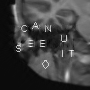} \\
    
    $\qquad \quad \text{Original } x_3 + r_3$ & $\, \, \, \text{Original } x_2 + r_2 \, \text{(zoomed)}$ & $\quad \,\, \text{MRI-VN} \, f_3(A_3x_3 + r_3)$& 
    $\quad \,\, \text{SoA from } A_3(x_3+r_3)$\\
    \includegraphics[width=0.227\textwidth]{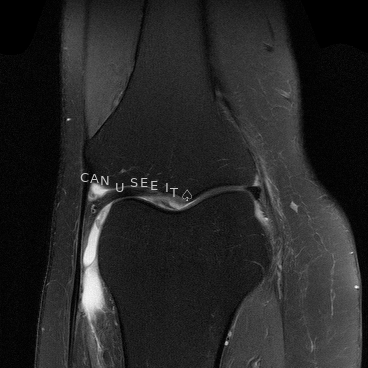} 
    &\includegraphics[width=0.227\textwidth]{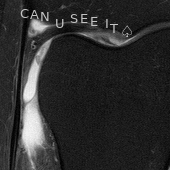} 
    &\includegraphics[width=0.23\textwidth]{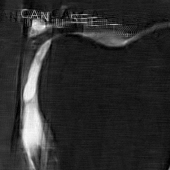} 
    &\includegraphics[width=0.23\textwidth]{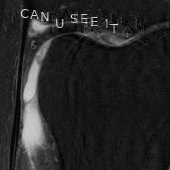} \\
    
     $\qquad \quad \text{Original } x_4 + r_4$ & $\, \, \, \text{Original } x_4 + r_4 \, \text{(zoomed)}$ & $\quad  \text{Deep MRI} \, f_4(A_4x_4 + r_4)$& 
     $\quad \,\, \text{SoA from } A_4(x_4+r_4)$\\
     \includegraphics[width=0.227\textwidth]{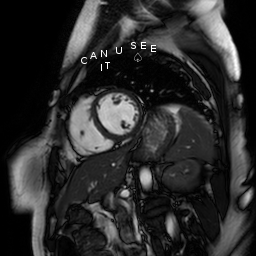} 
      &\includegraphics[width=0.227\textwidth]{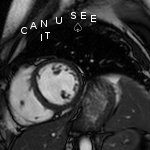}  
      &\includegraphics[width=0.23\textwidth]{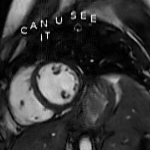} 
      &\includegraphics[width=0.23\textwidth]{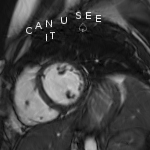} \\      
    
    $\quad\quad\quad \,$   Ell 50/Med 50 
    & $\quad\quad\quad\quad \,\,$ DAGAN 
    & $\quad\quad\quad\quad \,\,$ VN-MRI  
    & $\quad\quad\quad\quad \,\,$  Deep MRI 
    \\
    \includegraphics[width=0.23\textwidth]{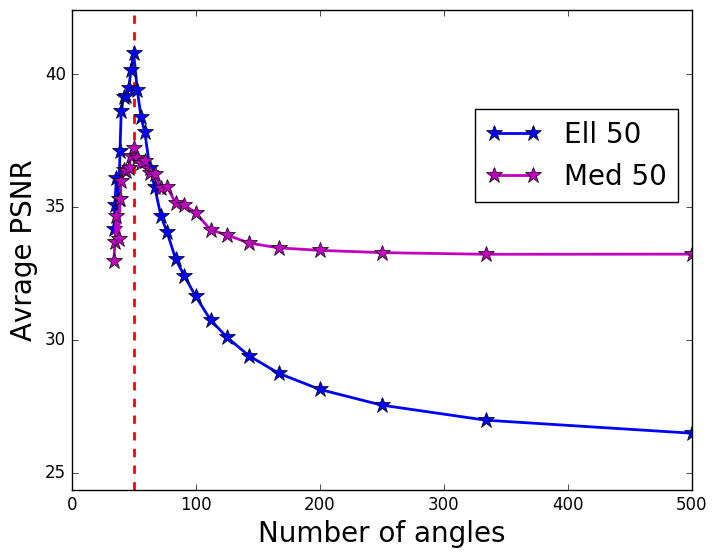}
  &\includegraphics[width=0.227\textwidth]{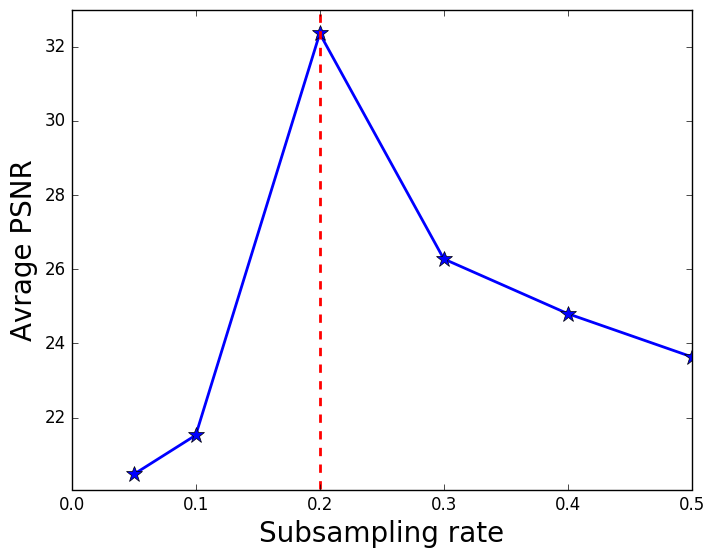} 
 &\includegraphics[width=0.23\textwidth]{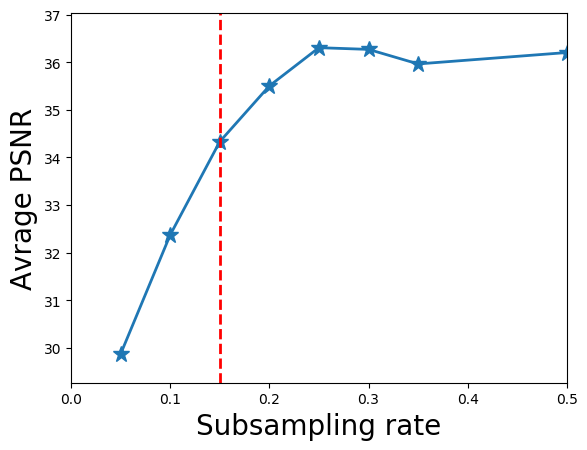} 
 &\includegraphics[width=0.227\textwidth]{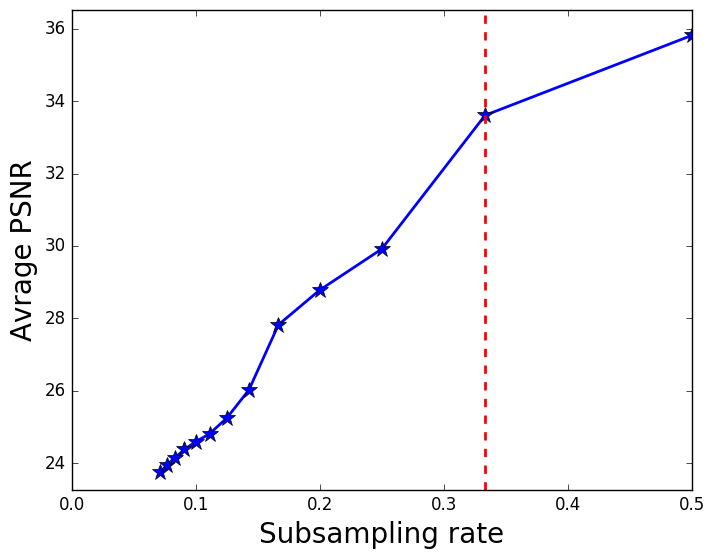} 
\end{tabular}
\caption{First four rows: Images $x_j$ plus structured perturbations $r_j$ (in the form of text and symbols) are reconstructed from measurements $y_j = A_j(x_j+r_j)$ with neural networks $f_j$  and state-of-the-art (SoA) methods. The networks are: $f_1= \text{Ell 50}$, $f_2 = \text{DAGAN}$, $f_3 = \text{MRI-VN}$, $f_4 = \text{Deep MRI}$. The sampling modalities $A_j$ are: $A_1$ is a subsampled discrete Radon transform, $A_2$ is a subsampled discrete Fourier transform (single coil simulation), $A_3$ is a superposition of subsampled discrete Fourier transform (parallel MRI simulation with 15 coils elements), $A_4$ is a subsampled discrete Fourier transform. Note that Deep MRI has not been trained with images containing any of the letters or symbols used in the perturbation, yet it is completely stable with respect to the structural changes. The same is true for the AUTOMAP network (see first column of Figure \ref{fig:automap_pert}). Last row: The figures show PSNR as a function of the subsampling rate for different networks. The red line indicates the subsampling ratio that the networks were trained for.}
  \label{fig:deep_mri_pert44} 
\end{figure*}

{\it Deep MRI:}
In this test we perturb the image $x$ with a sequence of perturbations  $\{r_j\}_{j=1}^3$ with $|r_1| < |r_2| < |r_3|$ in order to simulate how the instabilities continuously transform the reconstructed image from a very high quality reconstruction to an almost unrecognisable distortion. This is illustrated in the lower row of Figure \ref{fig:deep_mri_pert11}. Note that the perturbations are almost invisible to the human eye as demonstrated in the upper row of Figure \ref{fig:deep_mri_pert11}. The $r_j$ perturbations are created by early stopping of the algorithm iterating to solve for the optimal worst case perturbation. The purpose of this experiment is to demonstrate how the gradual change in perturbation create artefacts that may be hard to verify as non-physical. Indeed, the worst case perturbation $r_3$ causes clearly a reconstruction that, in a real world situation, can be dismissed by a clinician as non-physical. However, for the smallest $r_1$ we have a perturbation that is completely invisible to the human eye, yet it results in a reconstruction that is hard to dismiss as non-physical, and provides an incorrect representation of the actual image. Such examples could potentially lead to incorrect medical diagnosis. Note that state-of-the-art methods are not affected by the perturbation as demonstrated in the rightmost column of Figure  \ref{fig:deep_mri_pert11}. However, although this network is highly unstable with respect to tiny perturbations, it is highly stable with respect to small structured changes, see the 4th row of Figure \ref{fig:deep_mri_pert44}.

{\it AUTOMAP:} This experiment is similar to the one above, however, in this case we add $\tilde r_1, \hdots, \tilde r_4$ to the measurements $y = Ax$, where $|\tilde r_1| < |\tilde r_2| < |\tilde r_3| < |\tilde r_4|$ and $A$ is a subsampled discrete Fourier transform. In order to illustrate how small the perturbations are we have visualised $|x+r_j|$ in the first row of Figure \ref{fig:automap_pert}, where $y+\tilde r_j = A(x+r_j)$. To emphasise how the network reconstruction completely deforms the image we have, inspired by the second test on structural changes, added a small structural change in form of a heart that gradually disappears completely in the network reconstruction. This is demonstrated in the second row of Figure \ref{fig:automap_pert}, and the third row of Figure \ref{fig:automap_pert} contains the reconstruction done by a state-of-the-art method. Note that the worst case perturbations are completely different to the ones failing the Deep MRI network. Hence, the artefacts are also completely different. These perturbations are white-noise like and the reconstructions from the network provide a similar impression. As this is a standard artefact in MRI, it is, however, not clear how to protect against the potential bad tiny noise. Indeed, a detail may be washed out, as shown in the experiment (note the heart inserted with slightly different intensities in the brain image), but the similarity between the standard artefact may make it difficult to judge that this is an untrustworthy image. 

{\it MRI-VN:} In this case we add one perturbation $r_1$ to the image, where $r_1$ is produced by letting the algorithm searching for the worst perturbation run until it has converged. The results are shown in the first two columns of Figure \ref{fig:deep_mri_pert2}, and the conclusion is the same for the MRI-VN net as for Deep MRI and AUTOPMAP; perturbations barely visible to the human eye, even when zooming in, yield substantial misleading artefacts. Note also that  
the perturbation has no effect on the state-of-the-art-method.

{\it Med-50:} Here we add a perturbation $r_2$ that is also produced by running the algorithm until it has converged, and the results are visualised in the last two columns of Figure \ref{fig:deep_mri_pert2}. The Med-50 network is moderately unstable with respect to tiny perturbations compared to Deep MRI, AUTOMAP and MRI-VN, however, severe artefacts are clearly seen.
 It is worth noting that this network is used for the Radon transform, which is, from a stability point of view, a more unstable operator than the Fourier transform when considering its inverse.

\vspace{-2mm}

\subsubsection*{Stability with respect to small structural changes}

\vspace{-2mm}

Instabilities with respect to small structural changes are documented below.

{\it Ell-50:} This network provides a stark example of instability with respect to structural perturbation. Indeed, none of the details are visible in the reconstruction as documented in the first row of Figure \ref{fig:deep_mri_pert44} . 

{\it DAGAN:} This network is not as unstable as the Ell-50 network with respect to structural changes. However, as seen in the second row of Figure \ref{fig:deep_mri_pert44} the blurring of the structural details are substantial, and the instability is still critical. 

{\it MRI-VN:} This is an example of a moderately unstable network when considering structural changes. Note, however, how the instability coincides with the lack of ability to reconstruct details in general. This is documented in the third row of Figure \ref{fig:deep_mri_pert44}. 

{\it Deep MRI:} To demonstrate how the stability with respect to small structured changes coincides with the ability to reconstruct details, we show how stable the Deep MRI network is. Observe also how well the details in the image are preserved in the fourth row of Figure \ref{fig:deep_mri_pert44}. Here we have lowered the subsampling ration to $25\%$ even when the network is trained on $33\%$ subsampling ratio. We want to point out that none of the symbols, nor any text, has been used in the training set. 
\vspace{-2mm}

\subsubsection*{Stability with respect to more samples}

\vspace{-2mm}
Certain convolutional neural networks will allow for the flexibility of changing the amount of sampling. In our test cases all of the networks except AUTOMAP have this feature, and we report on the stability with respect to changes in the amount of samples below and in the last row of Figure \ref{fig:deep_mri_pert44}:

{\it Ell 50/Med 50:} 
 Ell 50 has the strongest and most fascinating decay in performance as a function of an increasing subsampling ratio. Med 50 is similar, however, with a less steep decline in reconstruction quality. 
 
 {\it DAGAN:} The reconstruction quality deteriorates with more samples similar to the Ell 50/Med 50 networks. 
 
 {\it VN-MRI:} This network provides reconstructions where the quality stagnates with more samples as opposed to the decay in performance witnessed in the other cases. 
 
 {\it Deep MRI:}
This network is the only one that behaves aligned with standard state-of-the-art methods and provides better reconstructions when more samples are added.

\vspace{-2mm}

\subsubsection*{Conclusion}

\vspace{-2mm}

The main conclusion of this paper is that the new paradigm of learning the reconstruction algorithm for image reconstruction in inverse problem, through deep learning, typically yields unstable methods.
 Moreover, our test reveals numerous instability phenomena, challenges and new research directions. In particular:

{\it (1) Tiny perturbations lead to a myriad of different artefacts.} Different networks yield different artefacts and instabilities, and as Figures \ref{fig:deep_mri_pert11}, \ref{fig:automap_pert}, \ref{fig:deep_mri_pert2} reveal there is no common denominator. Moreover, the artefacts may be difficult to detect as non-physical. Thus, several key questions emerge: given a trained neural network, which types of artefacts may the network produce? How is the instability related to the network architecture, training set and also subsampling patterns? 

{\it (2) Variety in failure of recovering structural changes.} There is a great variety in the instabilities with respect to structural changes as demonstrated in Figure \ref{fig:deep_mri_pert2}, ranging from complete removal of details to more subtle distortions and blurring of the features. How is this related to the network architecture and training set? Moreover, does the subsampling pattern play a role? 
It is important, however, to observe (as in the 4th row of Figure \ref{fig:deep_mri_pert44} and 1st column of  Figure \ref{fig:automap_pert}) that there are perfectly stable networks with respect to structural changes, even when the training set does not contain any images with such details. 

{\it (3) Networks must be retrained on any subsampling pattern.} The fact that more samples may cause the quality of reconstruction to either deteriorate or stagnate means that each network has to be retrained on every specific subsampling pattern, subsampling ratio and dimensions used. Hence, one may in practice need hundreds of different network to facilitate the many different combinations of dimensions, subsampling ratios and sampling patterns. 

{\it (4) Universality - Instabilities regardless of architecture?}
We have deliberately chosen networks with rather different architectures. Although our sample size of network is too small to conclude, it seems that instabilities may happen regardless of choice of architecture. However, different architectures may yield very different instabilities. 

{\it (5) Rare events? - Empirical tests are needed.} As misleading artefacts caused by instabilities may or may not be rare events, a vast amount of empirical statistical testing is needed in order to establish safe use of deep learning methods in image reconstructions.  However, such tests will have to be carried out on each specific network created on each specific subsampling pattern and dimension, yielding a potential vast research effort.  Moreover, the search for effective neural networks for inverse problems must take into account the complex instability phenomena raised here, not only generalisation.

{\footnotesize
\printbibliography[segment=1]}
\newrefsegment

\noindent\textbf{Supplementary Information} is included in our submission.\\
\\
\noindent\textbf{Acknowledgments}	
The authors would like to thank Dr. Cynthia McCollough, the Mayo Clinic, the
American Association of Physicists in Medicine, and the National Institute of
Biomedical Imaging and Bioengineering for allowing the use of their data in the
experiments.  F. Renna acknowledges support from the European Union's Horizon
2020 research and innovation programme under the Marie Sklodowska-Curie grant
agreement No 655282 and the FCT grant SFRH/BPD/118714/2016. B. Adcock
acknowledges support from the NSERC grant 611675.  ACH thanks NVIDIA for a GPU
grant in form of a Titan X Pascal and acknowledges support
from a Royal
Society University Research Fellowship, the UK Engineering and Physical
Sciences
Research Council (EPSRC) grant EP/L003457/1 and a Leverhulme Prize
2017.
\\
\\
\noindent\textbf{Correspondence} Correspondence should be addressed to A. Hansen (ach70@cam.ac.uk).

\noindent\textbf{Code and data} All code available from
\url{https://github.com/vegarant/Invfool}. Parts of the data is not publicly
available, but can be obtained by contacting the owner of the data. 

\newpage

\subsection*{Methods}
\begin{figure*}[t]
   \includegraphics[width=0.32\textwidth]{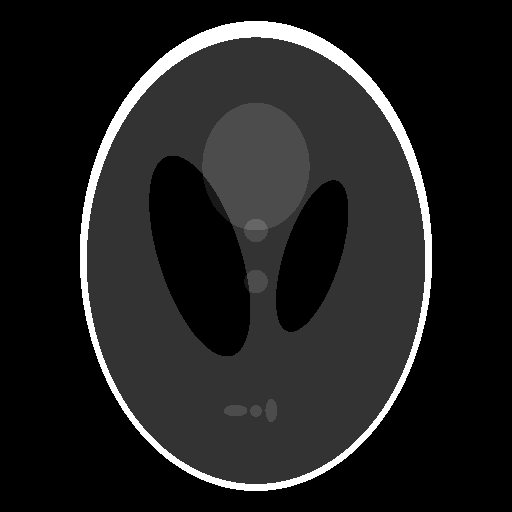} 
   \includegraphics[width=0.32\textwidth]{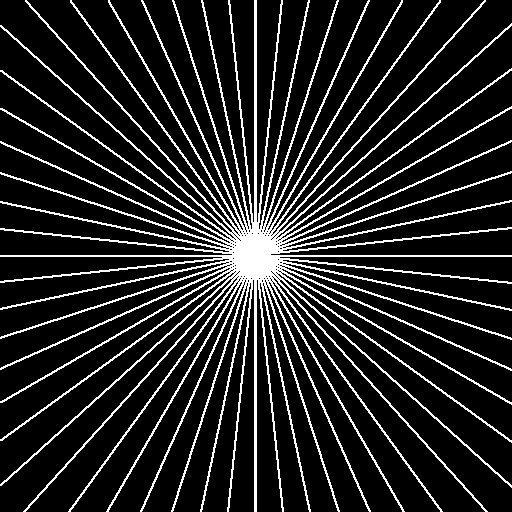} 
   \includegraphics[width=0.32\textwidth]{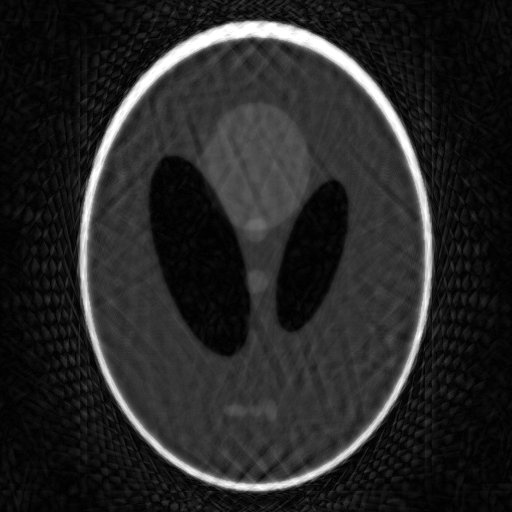} 
   \caption{ \label{fig:Fourier_adj_ex}
   (Under-sampled MRI problem) $x$ is the original image (left). $\Omega$ is the set of indices corresponding to the subsampling (middle). $\tilde x = A^*y$ (right) is the poorly reconstructed image, where $y = Ax$, $A = P_{\Omega}F$ and $F$ is the 2-dimensional discrete Fourier transform transform. Here $P_{\Omega}$ is the projection onto the span of $\{e_j\}_{j\in\Omega}$ where the $e_j$s denote the canonical basis.
 }
\end{figure*}

\begin{figure*}[t]
   \includegraphics[width=0.32\textwidth]{phantom.png} 
   \includegraphics[width=0.32\textwidth]{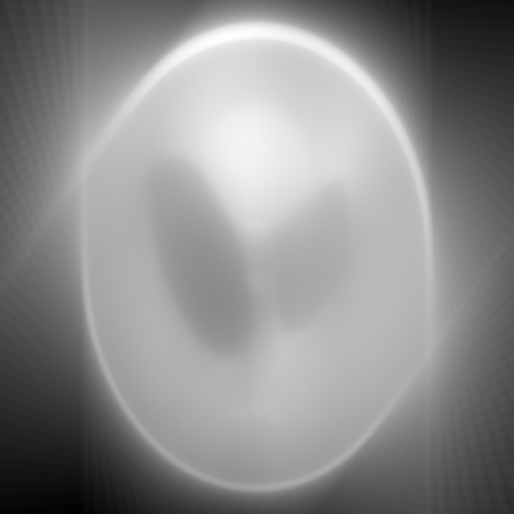} 
   \includegraphics[width=0.32\textwidth]{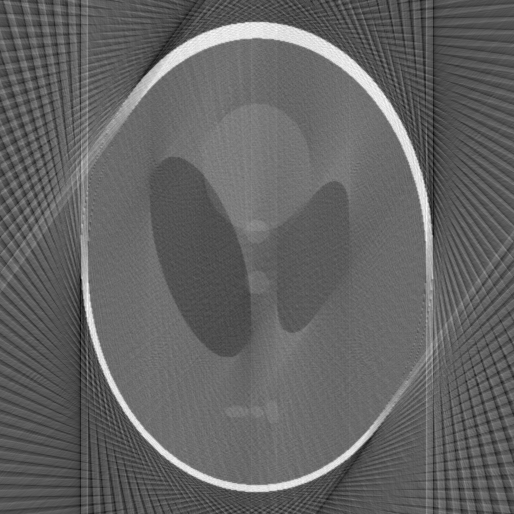} 
   \caption{ \label{fig:FBP_ex}
   (Under-sampled CT problem)  $x$ is the original image (left). $\hat x_1$ (middle) is the very poorly reconstruction, where $\hat x_1 = A^*y$, $y = Ax$ and $A$ is a subsampled discrete Radon transform. $\hat x_2$  (right) is the poorly reconstructed image, where $\hat x_2 = By$ and $B$ is a discrete filtered filtered back projection (FBP).}
\end{figure*}

The specific setups for deep learning and neural networks in inverse problems are typically rather specialised for each type of network. However, the main idea can be presented in a rather general way. Given an under-sampled inverse problem 
\begin{equation}\label{eq:inverse_prob}
Ax=y, \qquad A \in \mathbb{C}^{m \times N}, \qquad m < N
\end{equation}
there is typically an easy linear way of approximating $x$ from the measurement vector $y$. We will denote this linear operator by $H \in \mathbb{C}^{N \times m}$. In the MRI case, when $A$ is a subsampled discrete Fourier transform, often $H = A^*$. Note that in the MRI case $x$ is complex valued and we actually display the magnitude image $|x|$. An example is illustrated in Figure \ref{fig:Fourier_adj_ex}. In the CT case $H$ could be $A^*$, however, this gives very poor results (as demonstrated in Figure \ref{fig:FBP_ex}), and thus $H$ is usually a discretisation of the filtered back projection (FBP). The problem is, as displayed in Figure \ref{fig:FBP_ex}, that the reconstruction $\tilde x = Hy$ may still be rather poor. The philosophy of deep learning is quite simple; improve this reconstruction by using learning. In particular, inspired by deep learning in image denoising \cite{BurgerSH2012}, given training images $\{x_1,\hdots, x_n\}$ and poor reconstructions $\{HAx_1, \hdots, HAx_n\}$,
train a neural network $f: \mathbb{C}^N \rightarrow \mathbb{C}^N$ such that  
\begin{equation}\label{eq:phil_deep}
\|f(HAx_j) - x_j\| \ll \|HAx_j-x_j\|, \quad j = 1,\hdots, n.
\end{equation}
The hope is that \eqref{eq:phil_deep} will hold for other images as well, not just the training examples $\{x_1,\hdots, x_n\}$.

The construction process of the neural network $f$ is typically done as follows. First one decides on a particular class (architecture) of neural networks $\mathcal{NN}$. Then one decides on a cost function $\mathrm{Cost}: \mathcal{NN} \times \mathbb{C}^N \times \mathbb{C}^m \times  \mathbb{C}^N \rightarrow \mathbb{R}$ and tries to solve the optimisation problem of finding   
\begin{equation}\label{eq:opt1}
f \in  \mathop{\mathrm{arg min}}_{h \in \mathcal{NN}} \sum_{j=1}^n \mathrm{Cost}(h, HAx_j,Ax_j,x_j). 
\end{equation}
The task of finding a good class $ \mathcal{NN}$ and a good cost function $\mathrm{Cost}$ is an engineering art on its own. All the networks we test, except for AUTOMAP, are trained with some form of a "warm start" in form of a linear operator $H$, however, AUTOMAP is based on directly solving the problem
\begin{equation}\label{eq:opt2}
f \in  \mathop{\mathrm{arg min}}_{h \in \mathcal{NN}} \sum_{j=1}^n \mathrm{Cost}(h, Ax_j,x_j),
\end{equation}
without any reference to the reconstructions $Hx_j$. We refer to SI for details regarding the training of the networks. Note that the instability phenomenon is independent of the choice of \eqref{eq:opt1} or \eqref{eq:opt2}, and the operator $H$ may be viewed as just adding a layer to the network. Thus, we will in general talk about a network $f$ that takes the measurements $y = Ax$ as input.

\vspace{-2mm}

\subsubsection*{Describing the test}

\vspace{-2mm}

Before describing the algorithm for creating the unstable perturbations, it is convenient to have a short review of the framework for establishing instabilities for neural networks for image classification. For a detailed review of such methods,
see \cite{Fawzi17} and the references therein. The basic idea is as follows. Let $g \colon \R^d \to [0,1]^C$ be an image
classification network with $C$ different classes, so that  $g(x)$ is a
$C$-dimensional vector containing the probabilities associated to the
different classes for a given input image $x$. Let $\hat{k}_{g} \colon \R^{d}
\to \{ 1,\ldots, C\}$
where $\hat{k}_{g}(x) = \argmax_{i} (g(x)_i)$ is the image classifier. For a
given norm $\|\cdot\|$ on $\R^d$, we can then define the optimal, meaning smallest,
unstable perturbation $r^* \in \R^d$, for an image $x \in \R^{d}$ as  
\begin{equation}
    \label{eq:def_adv_class}
 r^{*}(x) \in  \mathop{\mathrm{arg min}}_{r}  \|r\| \,\text{ subject to }\, \hat{k}_{g}(x+r) \neq \hat{k}_{g}(x),
\end{equation}
where we write $r^*(x)$ to indicate that this is a perturbation for the image $x$.

It is clear that one cannot apply the approach in \eqref{eq:def_adv_class} to the problem of finding instabilities of neural networks for the inverse problem. Indeed, \eqref{eq:def_adv_class} is designed for a decision problem a la "is there a cat in the image?". In inverse problems there is no decision problem but rather the following, slightly simplified issue:
\begin{equation}\label{eq:inverse_problem}
\text{Reconstruct } x \text{ from } y = Ax, \qquad A \in \mathbb{C}^{m \times N}.
\end{equation}
Thus, if we are given a neural network $f\colon \C^m \to \C^N$ designed to solve \eqref{eq:inverse_problem}, and we want to search for instabilities imitating \eqref{eq:def_adv_class}, we would end up with the problem of finding
\begin{equation}\label{eq:first_attempt}
 \hat r(x) \in \mathop{\mathrm{arg min}}_{r}  \|r\| \text{ subject to } \|f(y + Ar) - f(y)\| \geq \delta,
\end{equation} 
for some $\delta > 0$,
where $y = Ax$ for some $x$. Note that \eqref{eq:first_attempt} has a clear disadvantage in that it may be infeasible for different values of $\delta$. Hence, a slightly different, constrained Lasso inspired variant, may be worth considering;
\begin{equation}\label{eq:second_attempt}
\tilde r(x) \in \mathop{\mathrm{arg max}}_{r} \|f(y + Ar) - f(y)\|  \quad \text{ subject to } \quad  \|r\| \leq \theta,
\end{equation}
for some $\tau >0$. In the case of \eqref{eq:second_attempt} we do not have any issues regarding infeasibility. However, a third option could be an unconstrained Lasso inspired version of \eqref{eq:second_attempt} given by 
\begin{equation}
    \label{eq:net_lasso}
    r^*(y) \in \mathop{\mathrm{arg max}}_{r} \frac{1}{2}\|f(y + Ar) - f(y)\|_{2}^{2} - \frac{\lambda}{2} \|r\|_{2}^{2}
\end{equation}  
(here we have specified the norm), where $\lambda > 0$.
Note that \eqref{eq:net_lasso} is not the only possibility. In particular, one could consider the more general setting
\begin{equation}
    \label{eq:net_lasso2}
    r^*(y) \in \mathop{\mathrm{arg max}}_{r}  \frac{1}{2}\|f(y + Ar) - p(x)\|_{2}^{2} - \frac{\lambda}{2} \|r\|_{2}^{2},
\end{equation}  
where $p: \mathbb{C}^N \rightarrow \mathbb{C}^N$. In this case $r^*$ will obviously depend on $p$, and the quality of the artefacts produced by $r^*$ may differ greatly as $p$ changes.  Indeed, this is the motivation for allowing this extra variable. In this paper we focus on \eqref{eq:net_lasso2} and consider $p(x) = f(Ax)$ (as in \eqref{eq:net_lasso}) and $p(x) = x$. 

However, the first part of our test could indeed be carried out by a different optimisation problem. Moreover, we anticipate that there will be other methods for creating instabilities for neural networks for inverse problems that will be as reliable and diverse as what we present here. 
Note that \eqref{eq:net_lasso2} is set up to find perturbations in the image domain. We do this deliberately as this provides an easy way to compare the original image with a perturbed image and deduce whether the reconstruction of the perturbed image is acceptable/unacceptable. However, one could set up \eqref{eq:net_lasso2} so that the perturbation is in the sampling domain as well.  
In the following we describe the test in detail and the methodology. 
\vspace{-2mm}

\begin{figure*}[htbp]
    \centering
    \vspace{-1cm}
    \begin{tabular}{m{0.175\textwidth}m{0.175\textwidth}m{0.175\textwidth}m{0.175\textwidth}m{0.175\textwidth}}   
    $\qquad \quad \text{Original }  x$ & $\quad\quad\quad\quad |x + r_1|$ & $\quad\quad\quad\quad |x + r_2|$& $\quad\quad\quad\quad  |x + r_3|$& $\quad\quad\quad\quad   |x + r_4|$\\

     \includegraphics[width=0.19\textwidth]{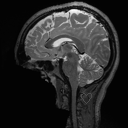}
    &\includegraphics[width=0.19\textwidth]{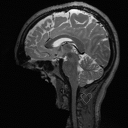}
    &\includegraphics[width=0.19\textwidth]{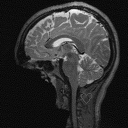}
    &\includegraphics[width=0.19\textwidth]{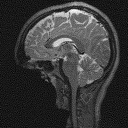}
    &\includegraphics[width=0.19\textwidth]{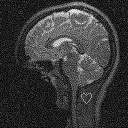}\\
 
  $\text{AUTOMAP (AM)} \, f(Ax)$ & $\quad \text{AM} \, f(A(x + r_1))$ & $\quad \text{AM}\, f(A(x + r_2))$ & $\quad \text{AM} \, f(A(x + r_3))$ & $\quad \text{AM} \, f(A(x + r_4))$\\

     \includegraphics[width=0.19\textwidth]{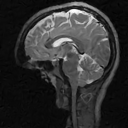}
    &\includegraphics[width=0.19\textwidth]{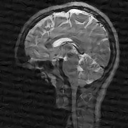}
    &\includegraphics[width=0.19\textwidth]{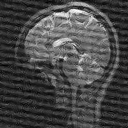}
    &\includegraphics[width=0.19\textwidth]{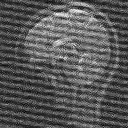}
    &\includegraphics[width=0.19\textwidth]{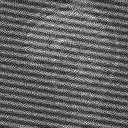}\\
    
      SoA from $A(x)$ &  SoA from $A(x+r_1)$ &  SoA from $A(x+r_2)$ &  SoA from $A(x+r_3)$ &  SoA from $A(x+r_4)$\\

     \includegraphics[width=0.19\textwidth]{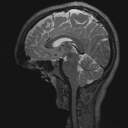}
    &\includegraphics[width=0.19\textwidth]{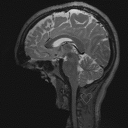}
    &\includegraphics[width=0.19\textwidth]{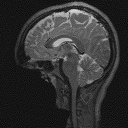}
    &\includegraphics[width=0.19\textwidth]{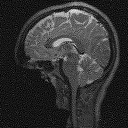}
    &\includegraphics[width=0.19\textwidth]{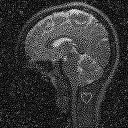}\\

    \end{tabular}
    \caption{\label{fig:automap_pert2} 
 The experiment from Figure \ref{fig:automap_pert} is repeated, however, by using a different $p$ in Algorithm 1. In particular, Figure \ref{fig:automap_pert} is produced by using $p(x) = x$, however, this figure is produced by using $p(x) = f(Ax)$. Note the substantial difference in the quality of the artefacts in the AUTOMAP reconstruction compared to Figure \ref{fig:automap_pert}.}
\vspace{-4mm}
\end{figure*}

\subsubsection*{Stability with respect to tiny perturbations}

\vspace{-2mm}

The neural network $f\colon \C^{m}\to \C^N$ is a non-linear function. In practice 
this makes the problem of finding a global maximum of the 
optimization problem in \eqref{eq:net_lasso2} impossible, even for small values
of $m$ and $N$.  In the following we will provide a method that aims at
locating local maxima of $\eqref{eq:net_lasso2}$ by using a gradient search
method. In particular, given an image $x \in \mathbb{R}^N$, $A \in \mathbb{C}^{m\times N}$ and $y = Ax$ as in \eqref{eq:inverse_prob} let 
\begin{equation}
\label{eq:obj_net_lasso}
Q^p_y(r) = \frac{1}{2}\|f(y+Ar) - p(x)\|_{2}^{2} - \frac{\la}{2} \|r\|_{2}^{2}  
\end{equation}
be the objective function. A most natural method to solve \eqref{eq:net_lasso2} is {\it gradient ascent with momentum}. Thus, the method uses the gradient of $Q^p_y$ in conjunction with two parameters $\gamma > 0$ (the momentum) and $\eta > 0$ (the learning rate)
in each step towards a local maximum. 

\begin{algorithm}[H]
\caption{\label{alg:invfool}Finding unstable perturbation for inverse problems}
    \begin{algorithmic}[1]
    \State $\textbf{Input: }\textit{Image: } x$,  $\textit{neural network: } f$, 
    $\textit{sampling matrix: } A$, $\textit{maximum number of iterations: } M$. 
    \State $\textbf{Output: } \textit{Perturbation } r_{M}$
    \State $\textbf{Initialize: }y \gets Ax$, $v \gets 0, ~i\gets 1, ~ r_0 \sim \text{Unif}([0,1]^N),$ $0< \lambda, \gamma,\eta, \tau$. Set $Q^p_y(r)$ as in Equation \eqref{eq:obj_net_lasso}.  
    \State $r_0 \gets \tau r_0$
    \While {$i \leq M$}
        \State $v_{i+1} \gets \gamma v_i + \eta \nabla_{r} Q_y(r_i)$
        \State $r_{i+1} \gets r_i + v_{i+1}$
        \State $i \gets i + 1$
    \EndWhile
    \State $\textbf{return } r_{M}$
    \end{algorithmic}
\end{algorithm}

This means that there are three parameters $\lambda > 0$, $\gamma > 0$ and $\eta > 0$ to be set, and hence the perturbation $r$ found by the algorithm will depend on these. 
The complete algorithm is presented in Algorithm
\ref{alg:invfool}, where $r_0$
is initialised randomly. Note that the parameter $\tau$ used in Algorithm
\ref{alg:invfool} is simply a scaling factor needed as the input images may have values in different ranges.

Note that for $u = y + Ar$, the gradient of $Q^p_y$ is given by 
\begin{equation}\label{eq:gradient}
 \nabla_{r}Q^p_y = A^* \nabla_{u}g(u) -  \lambda r, \quad g(u) := \|f(u)-p(x)\|_2^2
\end{equation}
where $\nabla_{u} g(u)$ can be computed efficiently using back
propagation. Note also that at each iteration this gradient is left multiplied by the
adjoint $A^*$. 

\begin{algorithm}[H]
\caption{\label{alg:invfool2}Finding unstable perturbation for Radon problems}
    \begin{algorithmic}[1]
    \State $\textbf{Input: }\textit{Image: } x$,  $\textit{neural network: } f$, 
    $\textit{sampling matrix: } A$, $\textit{FBP operator: } B$,  $\textit{maximum number of iterations: } M$. 
    \State $\textbf{Output: } \textit{Perturbation } r_{M}$
    \State $\textbf{Initialize: }y \gets Ax$, $v \gets 0, ~i\gets 1, ~ r_0 \sim \text{Unif}([0,1]^N),$ $0 < \lambda, \gamma,\eta, \tau$. Set $g(u)$ as in Equation \eqref{eq:gradient}.  
    \State  $r_0 \gets \tau r_0$
    \While {$i \leq M$}
        \State $v_{i+1} \gets \gamma v_i + \eta B \nabla_{u} g(y+Ar_i) - \lambda r_i$ 
        \State $r_{i+1} \gets r_i + v_{i+1}$
        \State $i \gets i + 1$
    \EndWhile
    \State $\textbf{return } r_{M}$
    \end{algorithmic}
\end{algorithm}

Just as in the case when training neural networks using stochastic gradient descent with momentum, choosing the parameters $\gamma$ and $\eta$ is an art of engineering. We are in a similar situation with our algorithm, and the optimal choices of $\gamma, \eta$ are based on empirical testing. 
Such experimenting with parameters also motivates experimenting with other parts of the algorithm. For example, when considering Radon measurements, we found that setting the optimal values of $\gamma$ and $\eta$ could be rather difficult. However, by replacing $A^*$ in \eqref{eq:gradient} by $B \in \mathbb{R}^{N \times m}$ being a discretisation of a Filtered Back Projection (FBP), this problem could be overcome and we therefore use Algorithm 2 in the case of Radon samples.

It should be mentioned that there are different variations of discretisations of the filtered backprojection for Radon problems. The discretisation $B \in \mathbb{R}^{N \times m}$ used in our experiment is the one provided by \textsc{MatLab} R2018b.

\begin{figure*}[t]
\centering
   \includegraphics[width=0.32\textwidth]{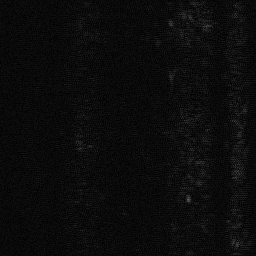} 
   \includegraphics[width=0.32\textwidth]{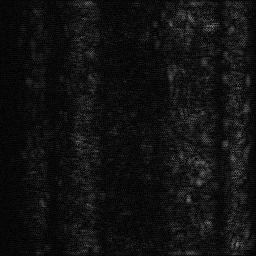} 
   \includegraphics[width=0.32\textwidth]{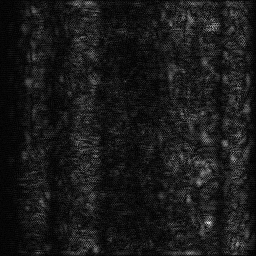} 
   \caption{\label{fig:pert1}
   We visualise the perturbations $|r_1|, |r_2|$ and $|r_3|$ used in Figure \ref{fig:deep_mri_pert11} and in the main manuscript to create the instabilities for the Deep MRI network. These perturbations have been rescaled to all lie in the same intensity range. The number of iterations in Algorithm \ref{alg:invfool} is given by
    $M=2000,4000,6000$, and the values of $\lambda,$ $\gamma$, $\eta$ and $\tau$ are given in Table \ref{tab:summary}.}
\end{figure*}

\begin{figure*}[t]
   \includegraphics[width=0.245\textwidth]{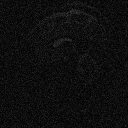} 
   \includegraphics[width=0.245\textwidth]{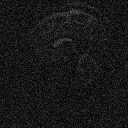} 
   \includegraphics[width=0.245\textwidth]{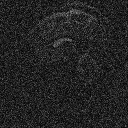} 
   \includegraphics[width=0.245\textwidth]{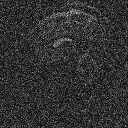} 
   
   \vspace{1mm}
   
   \includegraphics[width=0.245\textwidth]{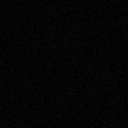} 
   \includegraphics[width=0.245\textwidth]{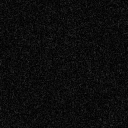} 
   \includegraphics[width=0.245\textwidth]{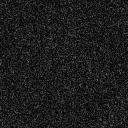} 
   \includegraphics[width=0.245\textwidth]{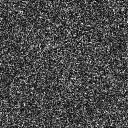} 
   \caption{\label{fig:pert2}
   We visualise the perturbations $|r_1|, |r_2|, |r_3|$ and $|r_4|$ used in
Figure \ref{fig:automap_pert} (top row) and Figure \ref{fig:automap_pert2}
(bottom row) to create the instabilities for the AUTOMAP network. These
perturbations have been rescaled to all lie in the same intensity range. The
number of iterations in Algorithm \ref{alg:invfool} is given by
    $M=12,16,20,24$, for the top row and $M=160,170,177,183$ for the bottom row. The values of $\lambda,$ $\gamma$, $\eta$ and $\tau$ are given in Table \ref{tab:summary}.}
\end{figure*}
\vspace{-2mm}

\subsubsection*{State-of-the-art comparison method}

\vspace{-2mm}

All of our tests are done against state-of-the-art benchmark methods using established techniques based on sparse regularisation and compressed sensing \cite{CandesRombergTao, donohoCS, Rudin, SMSparsity}.

There are many variations in the literature using X-lets and Total Variation
(TV) techniques separately or in combination.  Our main algorithm is based on
the re-weighting technique suggested in \cite{Candes_reweight}. This idea was
refined in \cite{Jackie_code_application} and \cite{Ma16}, by combining both X-lets (shearlets in this case) and TV. This is our main algorithm of choice used in this paper. We refer the reader to \cite{Ma16} for details, however, a
short summary can be described as follows. The algorithm allows for both
Fourier and Radon sampling, however, the current
implementation only allows for single coil MRI in the Fourier case. The
idea is to solve iteratively the problem 
\begin{equation*}
\minimize_{z} \sum_{j=1}^{J} \la_{j} \|W_{j}\Psi_{j} z\|_1 + \mathrm{TGV}_{\alpha}^{2}(z) \, \text{subject to} \, Az=y,
\end{equation*}
where the $\lambda_j$s are weights, $W_j$ is a diagonal weighting matrix,
$\Psi_j$ is the $j$'th subband in a shearlet transform \cite{Ma16}, and $\mathrm{TGV}_{\alpha}^2$, $\alpha = (\alpha_1,\alpha_2)$ is
the second order Total Generalised Variation operator. The $\mathrm{TGV}_{\alpha}^{2}$ operator
consist of a first order term (TV) weighted by $\alpha_1$ and a second order (generalised) term weighted by 
$\alpha_2$. In each iteration step the weights $\lambda_j$ and weighting
matrices $W_j$ are updated. 

In particular, the minimisation problem is casted into an unconstrained formulation 
\begin{equation*}
\minimize_{z} \sum_{j=1}^{J} \la_{j} \|W_{j}\Psi_{j} z\|_1 + \mathrm{TGV}_{\alpha}^{2}(z) + \frac{\beta}{2} \| Az -y \|_2^2,
\end{equation*}
and solved via split Bregman iterations. This means that the problem is decoupled into two portions, one accounting for the $\ell_1$-norm term and one for the $\ell_2$-norm term. 

In particular, on denoting by ${\Psi}'$ a composite operator including (1) the effect of multi-scale X-lets transform in different levels including the weights $\la_j$, 
(2) the first order (TV) term of $\mathrm{TGV}_{\alpha}^{2}$ and (3) the second order term of the same operator, and by adding a further splitting variable $d= \Psi' z$, it is possible to write the $k$-th split Bregman iteration as
\begin{equation*}
\left\{
\begin{array}{rcl}
(z_{k+1},d_{k+1}) &= & \arg \min_{z,d} \| W  d \|_1 + \frac{\beta}{2} \| Az - y_k  \|_2^2 \\
 & & + \frac{\mu}{2} \|  d - \Psi'z -b_k  \|_2^2, \\
 b_{k+1} & = & b_k + \Psi' z_{k+1} -d_{k+1}, \\
 y_{k+1} & = & y_k + y - A z_{k+1}.
\end{array}
\right.
\end{equation*}
 
During each iteration, the $(z,d)$-minimisation problem is solved using one or multiple non-linear block Gauss-Seidel iterations, which alternate between minimising $z$ and $d$. Also, in contrast with the re-weighting strategy originally presented in \cite{Candes_reweight}, the weights in $W$ are updated not only after convergence to the solution of the $\ell_1$ minimisation problem, but weight updates are incorporated in the split Bregman iterations.
 
In the above iterations we  have allowed for a slight abuse of notation. We are using
$\mu=(\mu_1, \mu_2, \mu_3)$ and split the sum $\|d - \Psi'z - b_k\|^{2}_{2}$,
into three separate parts, depending on which of the terms of $\Psi'$ they 
come from, and weight each partial sum separately with $\tfrac{\mu_1}{2},\tfrac{\mu_2}{2}$ and
$\tfrac{\mu_3}{2}$, respectively (see equation (15) in \cite{Ma16} for details).

This method has been used for reconstruction from Fourier and Radon measurements, using two different setups. For the case of Fourier measurements, discrete shearlets are generated with 3 scales and with directional parameters $[1\  2\  2]$ (see \cite{Jackie_code_application} and \cite{Ma16} for details). The optimisation parameters are set as follows:
\begin{itemize}
\item $(\mu_1,\mu_2,\mu_3)$ : $(5000, 10, 20)$,
\item $(\alpha_1,\alpha_2)$: $(1, 1)$,
\item $\beta$: $10^{5}$,
\item $\epsilon$: $10^{-5}$,
\end{itemize}
Where $\epsilon$ is a parameter which is added in the denominator, of the
updating rule, for the weights $W$, to avoid division by zero.  Similarly,
image reconstruction from Radon measurements are obtained by using shearlets
with 4 scales and directional parameters $[0\  0\  1\  1]$ and with the
following parameter setup:
\begin{itemize}
\item $(\mu_1,\mu_2,\mu_3)$ : $(500, 0, 0)$,
\item $(\alpha_1,\alpha_2)$: $(1, 0)$,
\item $\beta$: $50$,
\item $\epsilon$: $10^{-8}$,
\end{itemize}
Notice in particular that $\mu_3 = \alpha_2 = 0$, hence we are only using
shearlets and a TV term as our regularizes. In all setups, we run the algorithm until convergence, i.e. between 50 and 500 iterations. 

The above approach is used in all examples except for the tests using the MRI-VN network which is designed for parallel MRI. 
For this imaging modality we have the following reconstruction problem. 
Let $\Om \subseteq \{1, \ldots, N\}$, $|\Om|=m$ and $P_{\Om} \in \R^{m
\times N}$ be the projection operator onto the canonical basis i.e.\ $P_{\Om}x
= (x_i)_{i\in \Om}$. Let $F \in \C^{N \times N}$ be the discrete Fourier transform (DFT) matrix. The Fourier sampling matrix  can then
be written as $A_f= P_{\Om} F$, for a given sampling pattern $\Om$. 
In parallel Fourier imaging we receive information from multiple coils elements at the same time. 
This is modeled by introducing diagonal sensitivity matrices 
$S_1, \ldots S_{c} \in \C^{N\times N}$ which weight the measurements, based on 
the environmental conditions of the sensing problem.
The corresponding measurement matrix is then written as 
\[ A_{pf} = 
\begin{bmatrix} 
P_{\Om}F && \\ &\ddots & \\ &&P_{\Om}F
\end{bmatrix}\begin{bmatrix} S_{1} \\ \vdots \\ S_{c} \end{bmatrix} \in \C^{m'\times N}. \]
where $m' = cm$. Note that for this sampling operator we might have $m' > N$,
which means that the corresponding linear system may be overdetermined. 
Given an image $x \in \mathbb{R}^N$ we let $y = A_{pf}x$ and use the SPGL1 algorithm \cite{Berg08} for solving 
\[
 \minimize_{z} \|z\|_{1}  \quad\text{subject to}\quad \|A_{pf} \Psi^{-1}z - y\|_{2} \leq \delta,
 \] 
where $\Psi \in \R^{N\times N}$ is the wavelet transform corresponding to the periodised Daubechies 2 wavelet with 3 levels.
In all the experiments we set $\delta = 0.01$.

\vspace{-2mm}

\subsubsection*{Non-uniqueness of the test - parameter dependency}

\vspace{-2mm}
Note that the test we provide can never become unique. Indeed, we choose to solve \eqref{eq:net_lasso2} with different choices of $p$, however, 
\eqref{eq:first_attempt} and \eqref{eq:second_attempt} could also be viable
alternatives. Moreover, all of these approaches depend on parameters $\delta$,
$\theta$ and $\lambda$ that have to be specified, and different values give
different worst-case perturbations. In addition, Algorithm \ref{alg:invfool}
and Algorithm \ref{alg:invfool2} designed to solve \eqref{eq:net_lasso2} depend
on the parameters $\gamma$, $\eta$ and $\tau$. Moreover, note also that there is no
built-in halting criteria in  Algorithm \ref{alg:invfool} and Algorithm
\ref{alg:invfool2}, but rather the parameter $M$ controlling the number of
iterations.  Thus, the stability test can never become a unique test, but
instead a collection of algorithms depending on different parameters. Hence, an
appropriate use of the test means running Algorithm \ref{alg:invfool} and
Algorithm \ref{alg:invfool2} varying the parameters. This is also what is done
in this paper, however, only the results based on the final parameters are
displayed in the figures. The final parameters chosen are listed in Table
\ref{tab:summary}.

\begin{table}[H]
\begin{center}
 \begin{tabular}{|l|c|c|c|c|c|}
        \hline
 Neural Network    & $\lambda$ & $\gamma$ & $\eta$ & $\tau$ & $p(x)$\\ 
\hline\hline
Deep MRI & 0.001 & 0.9 & 0.01 & 0.01 & $f(Ax)$\\ \hline
AUTOMAP Fig. \ref{fig:automap_pert} & 0.1 & 0.9 & 0.001 & $10^{-5}$ & $f(Ax)$\\ \hline
AUTOMAP Fig. \ref{fig:automap_pert2} & 0.1 & 0.9 & 0.001 & $10^{-5}$ & $x$\\ \hline
MRI-VN  & 1 & 0.9 & 0.005 & 0.001& $f(Ax)$ \\ \hline
MED 50 & 20 & 0.9 & 0.005 & 0.005 & $f(Ax)$\\ \hline
\end{tabular}
    \caption{\label{tab:summary}Summary of the different choices of parameters leading to the results reported in Figures 1-3 and 7.}
    \end{center}
    \vspace{-4mm}
\end{table}

In Figure \ref{fig:pert1} and Figure \ref{fig:pert2}, we display different perturbations
$r_j$ produced by Algorithm \ref{alg:invfool} with different values of $M$ corresponding to the experiments shown in Figures 1-2 in the main
manuscripts. The values of $\lambda$, $\gamma$, $\eta$ and $\tau$ are as in Table
\ref{tab:summary}. Note the difference between the perturbations depending on
the network. As the perturbations are tiny, they have been enlarged in order to
get a visual impression.

\vspace{-2mm}

\subsubsection*{Stability with respect to small structural changes}
This part is fully explained in the main manuscript. However, we want to emphasise that the symbols used in the experiment are chosen in order to assure that networks can recover important details. These can of course be replaced by other symbols as long as the ability of an algorithm to reconstruct these symbols correlate with the ability to recover other small important details. 

{\it Important note:} In our examples, it is irrelevant whether the symbols have been included in the training set or not. In fact, both the AUTOMAP and the Deep MRI networks have no problem recovering the symbols, see the first column of Figure \ref{fig:automap_pert} and Figure \ref{fig:automap_pert2}, where a heart is artificially added, and the fourth row of Figure \ref{fig:deep_mri_pert44}. Indeed, none of these networks have been trained on images containing the symbols, yet they can perfectly well recover them. 

\vspace{-2mm}

\subsubsection*{Stability with respect to more samples}

All of the networks we have tested, except AUTOMAP, are convolutional
neural networks (CNNs), which means that the trained weights 
come from convolutional layers. This has the advantage of reducing the number
of parameters we need to learn, compared to fully connected layers (dense
matrices), and may for large image sizes be the only alternative. 
Moreover, these CNNs can easily be adapted to other subsampling patterns as explained below. Thus, one can easily apply a network that is trained on 25\% subsampling, say, to input that uses, for example, 35\% subsampling. The question is whether the quality of the reconstruction is kept when increasing the subsampling ratio. 
The reason for the flexibility of the CNNs in our test is that they all depend on the operator $H \in \C^{N\times m}$, as 
described in \eqref{eq:phil_deep}, by
considering it as a non-learnable first layer. As the $H$ is non-learnable,
this allows for flexibility in our choice of $m$, since we know how to
construct $H$ for various values of $m$. 

In the last row of Figure 4 in the
main manuscript we have varied the number of samples $m$ and 
measured the image quality of the networks reconstruction using the peak
signal-noise-ratio (PSNR) between the magnitude images of the original and the
reconstructed image.  Figure 4 shows all of the networks, except the AUTOMAP
network, which learns a mapping directly from the measurement domain without
using a non-learned layer $H$. Below follows the description of each of the experiments visualised in
the last row of Figure 4. 

\textit{Ell 50:} We created 25 sinograms of images containing ellipses similar to the data in the network's training and test set. The sinograms were created with 1000
uniformly spaced angles (views) using the formula for the Radon transform of an
ellipse. We then considered an acceleration factor $k \in \{2,\ldots,30\}$, by
subsampling every $k$-th line among the 1000 views. The FBP of the subsampled
sinogram was given to the network and the PSNR of the reconstruction was
computed against the FBP of all 1000 views.  

\textit{Med 50:} We used a test set, provided by the authors of \cite{jin17},
consisting of 25 CT images from the Mayo Clinic. These images were
synthetically sampled, using a the discrete Radon transform from \textsc{MatLab},  at the same angles as the Ell 50 network. The subsampled
sinograms were mapped back into the image domain using a FBP and reconstructed
using the network. The PSNR values were computed with the original image as
ground truth. 

\textit{DAGAN:} We used 20 MR images of brain tissue from the test set,
and subsampled these images using the 1D Gaussian sampling patterns provided by the authors
of \cite{DAGAN}. These patterns have been generated for subsampling rates $1\%$, $5\%$,
$10\%$, $20\%$, $30\%$, $40\%$ and $50\%$.

\textit{MRI-VN:} We used image data from the networks test set, and picked one
image slice from 10 different patients. The image data was subsampled with
uniformly spaced lines (center lines was always included), at subsampling rates
$5\%$, $10\%$, $15\%$, $20\%$, $25\%$, $30\%$, $35\%$, and $50\%$. The PSNR was
computed with the magnitude image of the fully sampled images as reference. 

\textit{Deep-MRI:} We used 30 image slices from one MRI scan, and
subsampled each slice using lines sampled according to a Gaussian distribution. Extra
caution was taken, so that all lines sampled at a low sampling rate, was
included at higher sampling rates. We sampled with an acceleration rate
$2,\ldots, 14$. 

It should be noted that measuring image quality is a delicate issue. We
point out that no comparison based on the last row of Figure \ref{fig:deep_mri_pert44} on the reconstruction quality should
be made between the networks, as the PSNR is unfit to measure image quality
between different types of images \cite{huynh2008scope}. However, we are only interested in the change in PSNR, as a function of subsampling percentage, for each specific network. 

\vspace{-2mm}

{\footnotesize
\printbibliography[segment=2]}

\end{multicols}

\newpage

\title{Supplementary Information}
\def\@thanks{}
\author{}

\begin{mytitlepage}
\maketitle
\end{mytitlepage}
\vspace{-3cm}

\newrefsegment

\section{Overview}
The Supplementary Information (SI) contains all the extra material on neural networks that is useful in order to understand and reproduce all the experiments done in the paper. In particular, the SI displays the variety of different architectures and training sets used in the various experiments. The neural networks considered are:

\begin{itemize}
 
\item[(i)]
 {\it AUTOMAP} \cite{Rosen_Nature}:  The AUTOMAP neural network we test is for low resolution single coil MRI with 60\% subsampling.  In the paper \cite{Rosen_Nature} one mentions 40\% subsampling, but this apparent discrepancy is simply due to different interpretation of the word subsampling. We use the traditional meaning in sampling theory referring to x\% subsampling as describing that the total amount of samples used are x\% of full sampling. The actual network used in our experiment is trained by the authors of \cite{Rosen_Nature} and provided through private communication. The details of the architecture and training data are summarised in \S \ref{sec:automap}.
 
 \item[(ii)] 
{\it DAGAN} \cite{DAGAN}: This network is for medium resolution single coil MRI with 20\% subsampling. The network weights are not available online, however, complete instructions on how to reproduce the network used in \cite{DAGAN} are accessible. Moreover, the advice from the authors (through private communication) of \cite{DAGAN} was to follow these guidelines to obtain the network. Thus, based on these instruction, we have retrained a network that reproduces the results in \cite{DAGAN}. The details of the training data and architecture are summarised in \S \ref{sec:DAGAN}.

\item[(iii)]
{\it Deep MRI} \cite{DeepMRI2d}: This neural network is for medium resolution single coil MRI with 33\% subsampling. 
The network used in our experiments is trained by the authors of \cite{DeepMRI2d}, can be found online and we summarise the details on training data and architecture in \S \ref{sec:deepMRI}.

\item[(iv)]
 {\it Ell 50} \cite{jin17}: Ell 50 is a network for CT or any Radon transform based inverse problem. The number 50 refers to the number of lines used in the sampling in the sinogram. The training of the network is done by the authors of  \cite{jin17}. The network can be obtained online, and all the details can be found in \S \ref{sec:FBPConvNet}.

 \item[(v)]
 {\it Med 50} \cite{jin17}: Med 50 has exactly the same architecture as Ell 50
and is used for CT, however the training is done on a different dataset. The network is
trained by the authors of \cite{jin17} and network weights have been obtained through private
communication. The details are summarised in \S \ref{sec:FBPConvNet}. 
 
 \item[(vi)]

{\it MRI-VN} \cite{Hammernik18}: This network is for medium to high resolution parallel
MRI with 15 coil elements and 15\% subsampling. In order to show a variety of subsampling
ratios we have trained this network on a smaller subsampling percentage than
what the authors of \cite{Hammernik18} originally (25\% and 33\%) did in their
paper. As we already have 33\%, and 20\%, we want a test on even lower
subsampling rates.  All the remaining parameters are kept as suggested in the code provided by the authors of \cite{Hammernik18}, except for the subsampling ratios and batch size
(due to memory limitations). All the details are documented in \S \ref{sec:MRI-VN}.

\end{itemize}

\section{Deep learning and neural networks for inverse problems}

The goal of deep learning in inverse problems is to learn a neural network $f \colon \C^m \to \C^N$  taking the
measurements $y = Ax$ (where $A$ is the matrix representing the sampling modality) as input and producing an approximation to $x$ as its output. Many of the networks considered in this work are not directly applied to the measurements $y$, but attempt to take advantage of the knowledge of the structure of the forward operator $A$ by first applying a transformation to the measurements $y$, in particular we obtain $\tilde{x} = H y$. The transformation $H$ represents the adjoint operator when the forward operator is defined in the Fourier domain and a discretised filtered back projection (FBP) operator for the case of Radon measurements. 

\subsection{AUTOMAP}\label{sec:automap}

\subsubsection{Network architecture}

The AUTOMAP network \cite{Rosen_Nature} is proposed for  
image reconstruction from Radon measurements, spatial non-Cartesian Fourier
sampling, misaligned Fourier sampling and undersampled Cartesian Fourier
samples.  In this work we have tested the network trained for image
reconstruction from undersampled Cartesian Fourier samples. In contrast with
the other networks considered in this work, the  AUTOMAP network provides a direct mapping
of the Fourier measurements to the image domain without applying the adjoint
operator $H$ as a first step. 

The authors of \cite{Rosen_Nature} have not made their code publicly available,
and the weights from their paper \cite{Rosen_Nature} had not been stored.
However, they kindly agreed to retrain their network for us and save the
weights. The network architecture they trained deviates slightly in some of
activation functions reported in their paper \cite{Rosen_Nature}, however, the
network was trained on the same data and sampling pattern. Below we describe
the network architecture we received. The training parameters and data, are
reported as in the paper  \cite{Rosen_Nature}.

\begin{figure}
    \centering
    \includegraphics[width=0.6\textwidth]{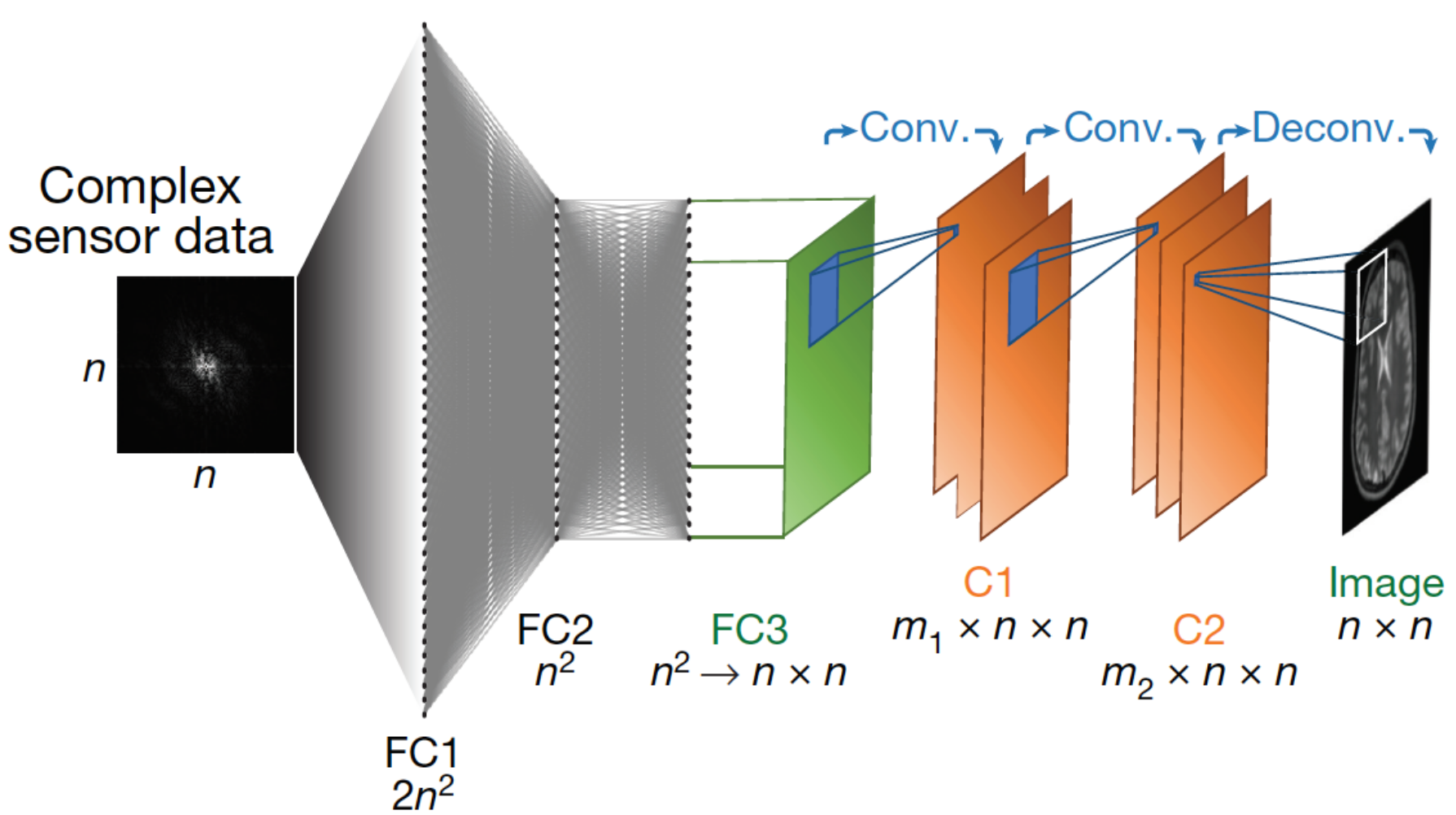} 
    \caption{\label{fig:AUTOMAP_arch}
The AUTOMAP architecture (figure from \cite{Rosen_Nature}).}
\end{figure}

The input of the AUTOMAP network, as described in \cite{Rosen_Nature} and in Figure \ref{fig:AUTOMAP_arch}   takes a complex $n \times n$ image of measurements as input. In the case of subsampling, one may interpret the $n \times n$ image as being zero padded in the coordinates that are not sampled. In the tests, $n = 128$, and in the actual implementation the input is represented by the complex measurement data
$y \in \mathbb{C}^{m}$ with $m=9855$  ($60\%$ of $n^2$) in this experiment. Such data is reshaped
into a vector of length $2 m $ with real entries before being fed into the network.
The first two layers of the network a fully connected matrices of size $25 000
\times 2m$ and $n^2 \times 25 000$. The first fully connected
layer is followed by a hyperbolic tangent activation function. The second 
fully connected layer is followed by a layer which subtracts the mean from the
output of the second layer. The output is then reshaped into an $n \times n$ image.

Next follows two convolutional layers with filter size $5\times 5$, 64 feature
maps and stride $1\times 1$.   The first convolutional layer is followed by a hyperbolic
tangent function, while the other is followed by a rectified linear unit
(ReLU).  Finally, the output layer deconvolves the 64 feature maps provided by
the second convolutional layer with $7 \times 7$ filters with stride $1\times 1$. The
output of the network is an $n \times n$ matrix representing the image
magnitudes.

\subsubsection{Training parameters}
The loss function used for training consisted of two terms,
$\mathcal{L}_{\text{SE}}$ and $\mathcal{L}_{\text{PEN}}$. Here
$\mathcal{L}_{SE}$ is the $\ell_2$-norm of the difference between the ground
truth magnitude image and the magnitude image predicted by the network.
Similarly the $\mathcal{L}_{\text{PEN}}$ is an $\ell_1$-penalty on the outputs
of the activations following the second convolutional layer ($C2$). The total
loss was then computed as 
\[ \mathcal{L}_{\text{TOTAL}} = \mathcal{L}_{\text{SE}} + \lambda \mathcal{L}_{\text{PEN}}\]  
with $\lambda = 0.0001$.
The network is trained using the RMSProp algorithm (see for example
\url{http://www.cs.toronto.edu/~
tijmen/csc321/slides/lecture_slides_lec6.pdf} as referred to in \cite{Rosen_Nature}) with minibatch size 100,
learning rate $0.00002$, momentum $0$, and decay $0.9$. The number of training
epochs is 100.

\subsubsection{Training data}
The training dataset consists of $50,000$ images taken from 131 different subjects from the MGH-USC HCP public dataset \cite{Fan16}\footnote{\url{https://
db.humanconnectome.org/}}. For each image, the central $256 \times 256$ pixels were cropped and subsampled to a resolution of $128 \times 128$ pixels. 
Before training, the images were preprocessed by normalizing the entire dataset to a constant value defined by the maximum intensity of the dataset.
Fourier data were obtained by subsampling the Cartesian $k$-space using a Poisson-disk sampling pattern with 60\% undersampling \cite{Uecker15}.

In order to increase the network robustness against translation, the following data augmentation scheme was applied. New images were created from each image in the training dataset by tiling together four reflections of the original image. Then, the so obtained $256 \times 256$ image was cropped to a random $128 \times 128$ selection. The routines used to implement the AUTOMAP network were written in 
TensorFlow\footnote{\url{https://www.tensorflow.org}}.

\subsection{DAGAN}\label{sec:DAGAN}
\subsubsection{Network architecture}
The DAGAN network was introduced in \cite{DAGAN} to recover images from Fourier
samples, with particular emphasis on MRI reconstruction applications. The DAGAN
network assumes measurements $y = Ax$, where $A$ is a subsampled discrete
Fourier transform. The input of the network is represented by the noisy
magnitude image $\tilde{x} = |Hy|$, which is obtained by direct inversion of
the zero-filled Fourier data, in particular, $H = A^*$.

The recovery algorithm presented in \cite{DAGAN} is based on a conditional
generative adversarial network (GAN) model, which consists of a generator
network, used for the image reconstruction, and a discriminator network,
measuring the quality of the reconstructed image. The generator network adopted
in \cite{DAGAN} has a U-net structure, similar to that used in \cite{jin17},
and its objective is to produce the recovered image. In \cite{DAGAN} the authors
propose two almost identical architectures, and train them with different loss
functions.  Below we will describe their ``refinement`` architecture trained with what
is referred to as Pixel-Frequency-Perceptual-GAN-Refinement loss in the paper.
The refined version is also our choice, as this architecture and training performed the best in the paper and in our tests as well. The network was not made publicly available, and based on advice from the authors of \cite{DAGAN} we trained the network ourselves.

\begin{figure}[t]
    \begin{center}
    \begin{tikzpicture}[scale=0.9]
        \draw[violet] (0, 0) -- (0, 6) node[above] {\textcolor{black}{1}};
\draw[yellow,->] (0.05, 3) -- (0.25, 3);
\node[above right] at (0,6) {$\gets$ number of filters};
\fill[violet] (0.3, 0.65625) rectangle (0.3448, 5.34375);
\node[above]  at (0.3224, 5.34375) {$64$};
\draw[red,->] (0.3948, 3) -- (0.5948, 3);
\fill[violet] (0.6448, 1.53516) rectangle (0.7344, 4.46484);
\node[above]  at (0.6896, 4.46484) {$128$};
\draw[red,->] (0.7844, 3) -- (0.9844, 3);
\fill[violet] (1.0344, 2.08447) rectangle (1.2136, 3.91553);
\node[above]  at (1.124, 3.91553) {$256$};
\draw[red,->] (1.2636, 3) -- (1.4636, 3);
\fill[violet] (1.5136, 2.4278) rectangle (1.872, 3.5722);
\node[above]  at (1.6928, 3.5722) {$512$};
\draw[red,->] (1.922, 3) -- (2.122, 3);
\fill[violet] (2.172, 2.64237) rectangle (2.5304, 3.35763);
\node[above]  at (2.3512, 3.35763) {$512$};
\draw[red,->] (2.5804, 3) -- (2.7804, 3);
\fill[violet] (2.8304, 2.77648) rectangle (3.1888, 3.22352);
\node[above]  at (3.0096, 3.22352) {$512$};
\draw[red,->] (3.2388, 3) -- (3.4388, 3);
\fill[violet] (3.4888, 2.8603) rectangle (3.8472, 3.1397);
\node[above]  at (3.668, 3.1397) {$512$};
\draw[red,->] (3.8972, 3) -- (4.0972, 3);
\fill[violet] (4.1472, 2.91269) rectangle (4.5056, 3.08731);
\node[above]  at (4.3264, 3.08731) {$512$};
\draw[blue,->] (4.5556, 3) -- (4.7556, 3);
\fill[green] (4.8056, 2.8603) rectangle (5.164, 3.1397);
\fill[violet] (5.164, 2.8603) rectangle (5.5224, 3.1397);
\node[above]  at (5.164, 3.1397) {$1024$};
\draw[blue,->] (5.5724, 3) -- (5.7724, 3);
\fill[green] (5.8224, 2.77648) rectangle (6.5392, 3.22352);
\fill[violet] (6.5392, 2.77648) rectangle (6.8976, 3.22352);
\node[above]  at (6.36, 3.22352) {$1536$};
\draw[blue,->] (6.9476, 3) -- (7.1476, 3);
\fill[green] (7.1976, 2.64237) rectangle (7.9144, 3.35763);
\fill[violet] (7.9144, 2.64237) rectangle (8.2728, 3.35763);
\node[above]  at (7.7352, 3.35763) {$1536$};
\draw[blue,->] (8.3228, 3) -- (8.5228, 3);
\fill[green] (8.5728, 2.4278) rectangle (9.2896, 3.5722);
\fill[violet] (9.2896, 2.4278) rectangle (9.648, 3.5722);
\node[above]  at (9.1104, 3.5722) {$1536$};
\draw[blue,->] (9.698, 3) -- (9.898, 3);
\fill[green] (9.948, 2.08447) rectangle (10.1272, 3.91553);
\fill[violet] (10.1272, 2.08447) rectangle (10.4856, 3.91553);
\node[above]  at (10.2168, 3.91553) {$768$};
\draw[blue,->] (10.5356, 3) -- (10.7356, 3);
\fill[green] (10.7856, 1.53516) rectangle (10.8752, 4.46484);
\fill[violet] (10.8752, 1.53516) rectangle (11.0544, 4.46484);
\node[above]  at (10.92, 4.46484) {$384$};
\draw[blue,->] (11.1044, 3) -- (11.3044, 3);
\fill[green] (11.3544, 0.65625) rectangle (11.3992, 5.34375);
\fill[violet] (11.3992, 0.65625) rectangle (11.4888, 5.34375);
\node[above]  at (11.4216, 5.34375) {$192$};
\draw[blue,->] (11.5388, 3) -- (11.7388, 3);
\draw[violet] (11.7888, 0) -- (11.7888, 6) node[above] {\textcolor{black}{64}};
\draw[teal,->] (11.8388, 3) -- (12.0388, 3);
\draw[violet] (12.0888, 0) -- (12.0888, 6) node[above] {\textcolor{black}{1}};
\draw[->, dashed] (0,0) -- (0,-0.6) -- (12.0888,-0.6) -- (12.0888, -0.45) node[above] {$+$};
\draw[orange,->] (12.1388, 3) -- (12.3388, 3);
\draw[violet] (12.3888, 0) -- (12.3888, 6) node[above] {\textcolor{black}{1}};
\draw[->, dashed] (0.3224,0.65625) -- (0.3224,0.35625) -- (11.444,0.35625) -- (11.444, 0.65625);
\draw[->, dashed] (0.6896,1.53516) -- (0.6896,1.23516) -- (10.9648,1.23516) -- (10.9648, 1.53516);
\draw[->, dashed] (1.124,2.08447) -- (1.124,1.78447) -- (10.3064,1.78447) -- (10.3064, 2.08447);
\draw[->, dashed] (1.6928,2.4278) -- (1.6928,2.1278) -- (9.4688,2.1278) -- (9.4688, 2.4278);
\draw[->, dashed] (2.3512,2.64237) -- (2.3512,2.34237) -- (8.0936,2.34237) -- (8.0936, 2.64237);
\draw[->, dashed] (3.0096,2.77648) -- (3.0096,2.47648) -- (6.7184,2.47648) -- (6.7184, 2.77648);
\draw[->, dashed] (3.668,2.8603) -- (3.668,2.5603) -- (5.3432,2.5603) -- (5.3432, 2.8603);
\draw[yellow, ->] (0, -0.85) -- (0.2,-0.85) node[right] {\textcolor{black}{$4\times4$ Conv2D stride (2,2)}};
\draw[red, ->] (0, -1.2) -- (0.2,-1.2) node[right] {\textcolor{black}{$4\times4$ Conv2D stride (2,2) +BN+ lReLU}};
\draw[blue, ->] (0, -1.55) -- (0.2,-1.55) node[right] {\textcolor{black}{$4\times4$ DeConv2D stride (2,2) +BN+ ReLU}};
\draw[teal, ->] (0, -1.9) -- (0.2,-1.9) node[right] {\textcolor{black}{$1\times 1$ Cond2D stride (1,1) +BN+ $\tanh$}};
\draw[orange, ->] (0, -2.25) -- (0.2,-2.25) node[right] {\textcolor{black}{ramp ($-1,1$)}};
    \end{tikzpicture}
    \end{center} 
    \caption{\label{fig:DAGAN_arch}DAGAN architecture. Here lReLU is the leaky ReLU function
with slope parameter equal to $0.2$.}
\end{figure}
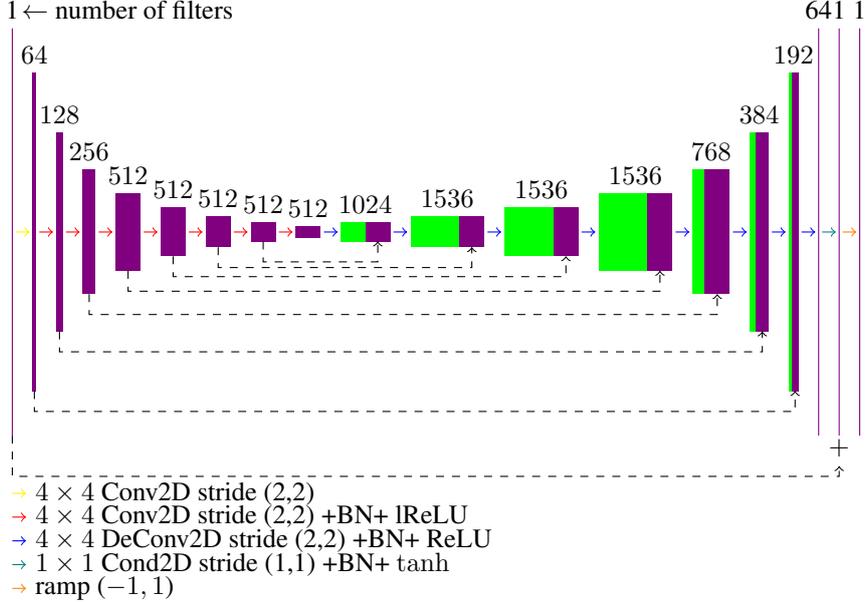

The architecture of the generator network, which is reported in Figure
\ref{fig:DAGAN_arch}, contains 8 convolutional layers and 8 deconvolutional
layers each followed by batch normalization (BN) \cite{Ioffe15}. The batch
normalization layers after the convolutional layers are followed by leaky ReLU
(lReLU) activations with slope equal to $0.2$ for $x <0$, while the batch
normalization layers after the deconvolutions are followed by a ReLU
activation. The generator network also contains skip connections, i.e.,
connections that copy the output of a layer directly to the input of a layer
further down in the hierarchy. The skip connections are used to concatenate
mirrored layers (see Figure \ref{fig:DAGAN_arch}).  The filter kernels used for
the convolutional and deconvolutional layers have size $4 \times 4$ with stride
$2 \times 2$. The number of filters in each convolutional/deconvolutional layer
increases/decreases according to Figure \ref{fig:DAGAN_arch}. 

The last deconvolutional layer is followed by a hyperbolic tangent activation
function.  A global skip connection, adding the input to the network
and the output from the hyperbolic tangent function, is then followed by
a ramp function clipping the output values of the image to the range
$[-1,1]$. Adding this last skip connection means that the network is actually
approximating the residual error between the network input $ \tilde{x} = Hy$
and the image of interest. 

The generator network is trained jointly with a discriminator network, which
aims to distinguish between the output of the generator network and ground
truth images. For further information on this network, we refer to \cite{DAGAN}.

\subsubsection{Training parameters}
The loss function used to train the DAGAN network consists of four different terms.
First, an image domain mean square error (MSE) loss, $\mathcal{L}_{\text{iMSE}}$, which accounts for the
$\ell_2$ distance between the output of the generator network and the ground
truth image. Second, a frequency domain MSE loss, $\mathcal{L}_{\text{fMSE}}$, which enforces consistency
between the output of the generative network in the frequency domain and the
acquired Fourier measurements. Third, a perceptual
loss term, $\mathcal{L}_{\text{VGG}}$, which is computed by using a pretrained VGG-16 described
in~\cite{VGG}. In particular, the VGG-16 network
was trained over the ImageNet dataset\footnote{\url{http://www.image-net.org/}}
and the output of its conv4 layer was used to compute the loss term by
considering the $\ell_2$-norm of the difference between the VGG-16 output
corresponding to the ground truth image and the generator network output.
Finally, the fourth term, $\mathcal{L}_{\text{GEN}}$ is computed using a cross
entropy loss on the output of the
discriminator network. Adding these four terms together gives us the loss
\[
\mathcal{L}_{\text{TOTAL}} = \alpha \mathcal{L}_{\text{iMSE}} 
                           + \beta \mathcal{L}_{\text{fMSE}} 
                           + \gamma \mathcal{L}_{\text{VGG}} 
                           + \tau \mathcal{L}_{\text{GEN}}, \qquad \alpha,\beta, \gamma, \tau > 0.
\]
We used the same values for $\alpha,\beta, \gamma$ and $\tau$ as in
\cite{DAGAN}, in particular, $\alpha=15$, $\beta = 0.1$, $\gamma = 0.0025$ and $\tau =
1$. The generator and the discriminator network were jointly trained by
alternating gradient optimization. In particular, the Adam \cite{Kingma14}
optimizer was adopted, with initial learning rate $0.0001$, momentum $0.5$, and
minibatch size $25$. The learning rate was halved every 5 epochs. 
We applied the same early stopping rule as given in their 
implementation\footnote{\url{https://github.com/nebulaV/DAGAN}}. This is based 
on measuring the $\mathcal{L}_{\text{iMSE}}$ loss between the training set 
and validation set. We used the early stopping number $10$. In total this 
resulted in $15$ epochs of training.

\subsubsection{Training data}
The DAGAN network was trained using data from a MICCAI 2013 grand challenge
dataset\footnote{\url{http://masiweb.vuse.vanderbilt.edu/workshop2013/index.php/}}.
We removed all images from the dataset where less than 10\% of the pixel
values were non-black. In total we therefore used 15912 images for training and
4977 images for validation. The dataset consisted of $\mathrm{T}_1$-weighted
MR images of different brain tissues.

The following data augmentation techniques were used to increase the amount of training data: image flipping, rotation, shifting, brightness adjustment, zooming, and elastic distortion~\cite{Simard03}.

 The discrete Fourier transform of the training images were subsampled using 1D
Gaussian masks, i.e., masks containing vertical lines of data in the
$k$-space randomly located over the image according to a Gaussian distribution.
In our tests we trained a network to do recovery from 20\% subsampling. 
The code used to implement DAGAN was written using the
TensorLayer\footnote{\url{http://tensorlayer.readthedocs.io}} wrapper and
TensorFlow, and was made publicly available by the authors of \cite{DAGAN}.

\subsection{DeepMRINet}
\label{sec:deepMRI}
\subsubsection{Network architecture}
The Deep MRI net is used to recover images from their subsampled Fourier
measurements. Its architecture is built up as a cascade of neural networks,
whose input is represented by the blurry image obtained by direct inversion of
the measurements, i.e., $\tilde{x}=H y$. The networks in the cascade are
convolutional neural networks (CNN) designed as follows
\[ CNN_{i}(\tilde{x})= \tilde{x} + C_{rec}^{(i)}\rho(C_{rec-1}^{(i)}\cdots \rho(C_{1}^{(i)} \tilde{x} + b_1^{(i)})\cdots+ b_{rec-1}^{(i)}) + b_{rec}^{(i)},
 \]
where $\rho(z)=\max\{0,z\}$ is the ReLU activation function, whereas $C_{k}^{(i)}$ and $b_{k}^{(i)}$represent trainable
convolutional operators and biases, respectively, for the $i$th network. These networks are then tied together and interleaved with \emph{data consistency layers (DC)}, which have the objective to promote consistency between the reconstructed images and the Fourier measurements. The DC layers are defined as 
\[ DC_{\la}(\tilde{x},y, \Om) = F^{-1} g_{\la}(F\tilde{x}, y,\Om), 
\quad\text{ where } \quad
g_{\la}(z, y, \Om) = \begin{cases} 
z_{k} &k \not\in\Om \\
\frac{z_{k}+\la y_k}{1+\la} & k \in \Om
\end{cases}.  \]   
Here $F$ represents the Fourier operator, and $\Om$ is the set of indices 
corresponding to the measurements acquired in the $k$-space. We point out that in
the limit $\lambda \to \infty$, the $g_{\lambda}$ function simplifies to $y_k$
if $k\in \Om$ and $z_k$ otherwise.   

In practice, a DC layer performs a weighted average of the Fourier coefficients
of the image obtained as the output of a CNN in the cascade and the true
samples $y$. The parameter $\lambda$ can either be trained or kept fixed. In \cite{DeepMRI2d},
it is not specified whether $\lambda$ is learned or not, however, from the 
code\footnote{\url{https://github.com/js3611/Deep-MRI-Reconstruction}} it is clear that $\lambda$ is chosen to be $\infty$. 

\begin{figure}[t]
    \centering
    \includegraphics[width=1\textwidth]{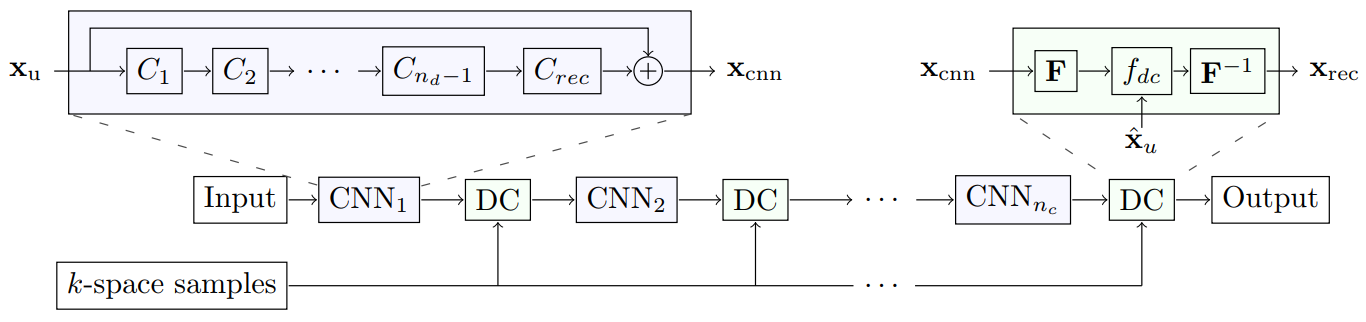} 
    \caption{\label{fig:deepMRINet_arch}
The DeepMRINet architecture (figure from~\cite{DeepMRI2d}).}
\end{figure}

  The complete network can now be written as 
\[ f(y,\Om) = DC_{\la}(CNN_{n}(\cdots DC_{\la}((CNN_1(H y)), y,\Om) \cdots),y, \Om), \]
and its architecture is reported in Figure \ref{fig:deepMRINet_arch}. In
particular, the architecture used to produce the results in \cite{DeepMRI2d}
and those reported in this paper contains 5 CNNs interleaved with 5 DC layers.
Each CNN contains 5 convolutional layers, all with kernel size $3\times 3$ and stride $1 \times 1$.
The first 4 layers are using 64 filters and are followed by a ReLU activation
function.  The fifth convolutional layer in each CNN contains 2 filters,
representing the real and imaginary part of the image. This fifth layer is not
followed by any activation function, however its output is added to
the input to the CNN using a skip connection. 

\subsubsection{Training parameters}
In our experiments we used a pre-trained network that was trained (and
published online) by the authors of \cite{DeepMRI2d} with training parameters
documented in the paper \cite{DeepMRI2d}.  The DeepMRINet was trained 
using a loss function with two terms, $\mathcal{L}_{\text{MSE}}$ and $\mathcal{L}_{\text{WEIGHTS}}$.
The $\mathcal{L}_{\text{MSE}}$ term computed the mean squared error (MSE) 
between the true (complex valued) 
image and the predicted (complex valued) image, while the 
$\mathcal{L}_{\text{WEIGHTS}}$ computed the $\ell_2$-norm of the weights.
The loss function was then computed as 
\[ \mathcal{L}_{\text{TOTAL}} = \mathcal{L}_{\text{MSE}} + 
10^{-7} \mathcal{L}_{\text{WEIGHTS}}.\]  
The network weights were initialized using He initialization \cite{he2015delving} and the Adam \cite{Kingma14}
optimizer was used for training. This optimizer takes as input a learning rate (step size) $\alpha$, and two 
exponential decay parameters $\beta_1$ and $\beta_2$ related to a momentum
term. We refer to \cite{Kingma14} for further explanations of these parameters.
The network was trained with  $\alpha = 10^{-4}$, $\beta_1=0.9$, $\beta_2=
0.999$  and batch size equal 10.

\subsubsection{Training data}
The DeepMRINet was trained using data from five subjects from the MR dataset
used in \cite{Caballero14}, which consists of 10 fully sampled short-axis
cardiac cine scans. Each of these scans was then preprocessed, using the SENSE
\cite{SENSE} software, into $30$ temporal (complex-valued) frames of size $256
\times 256$. Synthetic MRI measurements were then obtained by sampling
retrospectively the reconstructed images in $k$-space according to a Cartesian
undersampling masks. During training, whereas a fixed undersampling rate of
33\% was used, different undersampling masks were randomly generated in order
to allow the network to recover images from measurements obtained with
different undersampling masks. In particular, training images were fully
sampled along the frequency-encoding direction but undersampled in the
phase-encoding direction, according to the scheme described in \cite{Jung07}
(center frequencies were always included in the subsampling patterns).

To prevent overfitting, data augmentation was performed by including rigid
transformations of the considered images in the training datasets. 

The code used to implement the DeepMRINet was written in Python using the
Theano \footnote{\url{http://deeplearning.net/software/theano}} and
Lasagne\footnote{\url{https://lasagne.readthedocs.io/en/latest}} libaries.

\subsection{FBPConvNet -- The Ell 50 and Med 50 networks}
\label{sec:FBPConvNet}
The Ell 50 and Med 50 networks were proposed in \cite{jin17} under the name
FBPConvNet. The networks are trained to reconstruct images from Radon
measurements. The networks have identical architecture and are trained using
the same algorithm, with the same set of hyper parameters. The only difference
between the training of the two networks, is the dataset they have been trained
on. Below, we will describe the architecture and the training
procedure of both the networks. We will then describe the datasets
for the two networks in separate sections.
 
\subsubsection{Network architecture}
\begin{figure}
    \centering
    \includegraphics[width=0.8\textwidth]{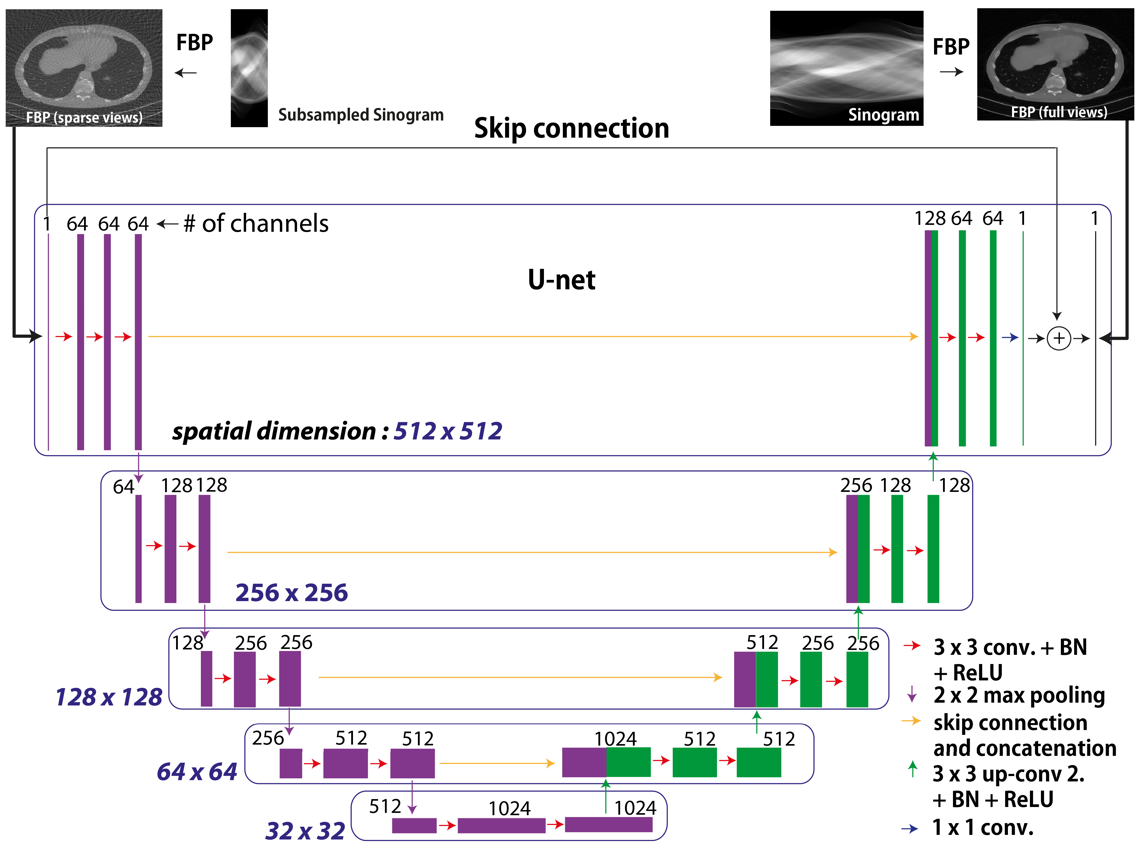} 
    \caption{\label{fig:FBPConvNet_arch}
The Ell 50 and Med 50 architecture (figure from \cite{jin17}).}
\end{figure}

The Ell 50 and Med 50 networks are trained for reconstructing $x$ from
measurements $y=Ax$ where $A \in \R^{m \times N}$  is a Radon\footnote{We used
\textsc{MatLab}s \lstinline{randon} command to represent this operator}
sampling operator, sampling 50 uniformly spaced radial lines. Rather than
learning a mapping from $y$ to $x$ directly, the networks takes advantage of a discrete filtered back projection\footnote{We used \textsc{MatLab}s
\lstinline{iradon} with linear interpolation and a `Ram-Lak` filter to represent this operator
} $H \in \R^{N \times m}$, as
described in the methods section, to obtain a noisy approximation $\tilde{x} =
Hy$ to $x$. The operator $H$ can be seen as a non-learnable first layer in the network.

The network contain several convolutional and deconvolutional layers, all of
which (except the last) are followed by a batch normalization (BN) layer and a ReLU
activation function. The (de)convolutional layers use filter size $3 \times 3$,
stride $1\times 1$ and a varying number of filters. We will not describe the
full architecture in detail, as it can be seen in Figure
\ref{fig:FBPConvNet_arch}, with the relevant number of filters, skip
connections, max-poolings ($2\times 2$) and concatenations. 
We do, however, point out that the network applies a final global skip
connection, so that rather than learning a mapping from $\tilde{x}$ to $x$
the network is trying to learn the ``noise" $x - \tilde{x}$.

\subsubsection{Training parameters}
The network weights were provided by the authors of \cite{jin17} and obtained 
based on the training procedure as described in their paper \cite{jin17}. The loss function
used to train the networks is the $\ell^2$ difference between the network
output and the ground truth, and the networks are trained using the stochastic
gradient descent algorithm with momentum. The learning rate varies from $0.01$ to $0.001$, whereas the momentum is set to $0.99$,
and the minibatch size is equal to 1.  During training, gradients are clipped
to the interval $[-I_{\mathrm{max}}, I_{\mathrm{max}}]$ with $I_{\mathrm{max}}
= 10^{-2}$, to prevent the divergence of the cost function. The networks are 
trained for 101 epochs, and the code used to implement the networks is written in 
\textsc{MatLab}  using the library MatConvNet\footnote{http://www.vlfeat.org/matconvnet}.

\subsubsection{Ell 50 -- Training data}

The Ell 50 network is trained from the filtered back projection of 475
synthetic sinograms containing the Radon transform of ellipses of random
intensity, size, and location.  The dynamic range of the back projected images is
adjusted so that image values are contained in the interval $[-500,500]$.
The Radon transform of an ellipse has an analytic formula, and hence this formula was
used to create sinograms of such images using $1000$ uniformly spaced lines
(views).
Measurement data are obtained by retaining $50$ radial lines out of the $1000$
views. The ground truth images were obtained by applying filtered back
projection to fully sampled sinograms (i.e., $1000$ radial lines). This
approach is motivated by the fact that in applications, one will never have
access to the underlying actual ground truth image. 
Data augmentation is also applied to the training data, by considering
horizontal and vertical mirroring of the original images.

\subsubsection{Med 50 -- Training data}
Med 50 is trained on synthetic images obtained from 475 real in-vivo CT images
from the Low-dose Grand challenge competition database provided by the Mayo
Clinic. The sinograms used for this training were synthetically generated from
high quality CT-images using \textsc{MatLab} \verb1radon1 command. The same approach
as for the Ell 50 network was used, where one sampled 1000 view and used this
as ground truth. The network was trained from 50 of these views. 

\subsection{MRI Variational Network (MRI-VN)}
\label{sec:MRI-VN}

\subsubsection{Network architecture}

The MRI Variational Network (MRI-VN) presented in \cite{Hammernik18} is
designed to reconstruct images from undersampled MRI data, sampled using
15 coil elements. Thus, we use the sampling operator $A = A_{pf}$ as described 
in the methods section, with $c=15$.

The network structure is inspired by the unfolding of a variational
minimization problem including a fidelity term and a regularization term
defined according the Fields of Experts model \cite{Roth09}. In particular,
each iteration of the corresponding Landweber method \cite{Landweber1951} corresponds to a layer of
the resulting neural network. More specifically, the implementation considered
in this work consists of $T=10$ layers/iterations that can be expressed as
follows:
\begin{equation}
\label{eq:vn_itr}
 u^{t+1} = u^{t} -   
(K^{t})^T \Psi^{t}(K^{t} u^{t}) 
+ \lambda^t A^{*}(A u^t - y), \quad 0\leq t< T 
\end{equation}
where $u^0 = Hy$ is the complex image obtained by applying $H=A_{pf}^*$. We will 
describe each of the remaining components of this network separately.

We start by noticing that the images $u^{t} \in \C^{N}$ (stacked as a vector in this simplified description) are complex valued, and can therefore described by its real and
imaginary components $u^{t}_{\text{re}}$ and $u^{t}_{\text{im}}$, respectively.
We will alternate between the representations.  

The operator $K^{t} \colon \C^{N} \to \R^{N\times N_{k}}$ acts as follows on $u^{t}$,
\[
K^{t}u^{t} = K_{\text{re}}^{t} u^{t}_{\text{re}} + K_{\text{im}}^{t} u^{t}_{\text{im}},
\]
where $K_{\text{re}}^{t}, K_{\text{im}}^{t} \colon \R^{N} \to \R^{N \times N_{k}}$, 
are learnable convolutional operators, with $N_{k}$ filters (channels), filter size $11 \times 11$ and 
stride $1 \times 1$. We will comment on the value of $N_{k}$ later. 

The $\Psi^t \colon \R^{N \times N_k} \to \R^{N \times N_{k}}$ is a non-linear
activation function in the network. For each filter/channel, $i=1,\ldots, N_k$
it applies the non-linear function
\[ 
\phi_{i}^{t}(z) 
= \sum_{j=1}^{N_w} w_{ij}^{t} \exp\left(-\frac{(z-\mu_{j})^2}{2\sigma^2}\right) ,
\] 
pointwise to each component $z$ in that channel. Here $\{w_{ij}^{t}\}_{i=1, j =
1}^{N_k, N_w}$, with $N_{w}=31$, are weights which are learnt during the
training phase. The nodes $\mu_j$ are non-learnable, and distributed in an
equidistant manner on the interval $[-I_{\mathrm{max}}, I_{\mathrm{max}}]$, for a
fixed value $I_{\mathrm{max}}$, commented on below. The $\sigma$ is also
non-learnable and equals $\frac{2 I_{\mathrm{max}}}{N_w-1}$.

The operator $(K^{t})^{T}\colon \R^{N \times N_{k}} \to \C^{N}$, maps
$z \mapsto (K^{t}_{\text{re}})^Tz + \mathrm{i} (K_{\text{im}}^{t})^{T}z$, where 
$\mathrm{i}$ is the imaginary unit, and $(K_{\text{re}}^{t})^T, (K_{\text{im}}^{t})^T$ are the transpose of $K_{\text{re}}^{t}, K_{\text{im}}^{t}$,  respectively.  
The matrices $A, A^*$ are the matrix $A_{pf}$ and its adjoint, while 
$\lambda^{t}$ is a learnable scalar. The remaining operations should be 
clear from Equation \eqref{eq:vn_itr}.

During training, each of the filters in $K_{\text{re}}^{t}$ and $K_{\text{im}}$
were restricted to have zero mean and
have unit Euclidean norm. This was done to avoid a scaling problem with the
weights $w_{ij}$.

To reproduce this network, we use the code published by the authors
of
\cite{Hammernik18}\footnote{
\url{https://github.com/VLOGroup/mri-variationalnetwork}}.  
Parts of this code uses slightly different parameters, than what was used
in the original paper. In particular, the value $N_{k}=24$ was chosen, rather
than $N_{k}=48$, as used in the paper. The value of $I_{\max}$, was
also changed from $150$ in the paper, to $1$ in the code.  The change of the
$I_{\max}$ value is motivated by another change, also made in the
published implementation, namely the scaling of the $k$-space values. In
\cite{Hammernik18} the $k$-space volumes (with $n_{sl}$ slices) was normalized
by the factor $\sqrt{n_{sl}}10000/\|y_{\text{volume}}\|_{2}$, whereas in the
code this have been changed to scaling each $k$-space slice $y$ with
$1/\|Hy\|_2$. This change has been made to make the their implementation
more streamlined. Whenever there has been a conflict between the two sources,
we have chosen the version found in the code.  

\subsubsection{Training parameters}
The MRI-VN network is trained using the $\ell_2$-norm as loss function. In particular, since MRI reconstruction are typically assessed through magnitude images, the error is evaluated by comparing smoothed version of magnitude images
\begin{equation*}
| x |_{\epsilon} = \sqrt{(\text{Re}(x))^2+(\text{Im}(x))^2 + \epsilon},
\end{equation*}
with $\epsilon = 10^{-12}$. 
The network parameters that minimize the loss function are determined using the
inertial incremental proximal gradient (IIPG) optimizer (see
\cite{Hammernik18,Kobler17} for details). Optimization is performed for $1000$
epochs, with a step size of $10^{-3}$. Training data is arranged into
minibatches of size 5. In the original paper, the batch size was set to 10, but
due to memory limitations we had to adjust this.

\subsubsection{Training data}

The authors in \cite{Hammernik18} considered 5 datasets for different types of
parallel MR imaging protocols, and trained one VN for each dataset. In this
work, we have trained a VN for one of these protocols, namely \emph{Coronal
Spin Density weighted with Fat Suppression}.  The training data consisted of 
knee images from 10 patients. From each patient we used 20 slices of the knee 
images, making up a total of 200 training images. The raw $k$-space data 
for each slice consisted of 15, $k$-space images of size $640 \times 368$, 
each with a precomputed sensitivity map. The sensitivity map was 
computed by the authors of \cite{Hammernik18}, using ESPIRiT \cite{espirit}.

The raw data was obtained using a clinical 3T system (Siemens Magnetom Skyra)
using an off-the-shelf 15-element knee coil. The raw data was subsampled
retrospectively by zeroing out 85\% of the $k$-space data. In 
\cite{Hammernik18} they test both a regular sampling scheme and a 
variable density pattern as proposed in \cite{Lustig07}. In our work, we used 
a regular sampling scheme, where the 28 first central $k$-space lines 
were sampled, and the remaining lines were placed equidistantly in $k$-space. 
No data augmentation was used.

The code was implemented in Python with a custom made version of 
Tensorflow\footnote{
\url{https://github.com/VLOGroup/tensorflow-icg}
}, 
which was partly implemented in C++/CUDA with cuDNN support. All the code and
data have been made available online by the authors of \cite{Hammernik18}.

{\footnotesize
\printbibliography[segment=3]}

\end{document}